\documentclass[runningheads]{llncs}
\usepackage{graphicx}
\usepackage{microtype}

\usepackage{latexsym}
\usepackage{setspace}

\usepackage{verbatim}
\usepackage{caption}
\usepackage{graphicx}
\usepackage{color}
\usepackage{multirow}
\usepackage{tabularx}
\makeatletter
\newcommand*{\addFileDependency}[1]{% argument=file name and extension
  \typeout{(#1)}
  \@addtofilelist{#1}
  \IfFileExists{#1}{}{\typeout{No file #1.}}
}
\makeatother

\newcommand*{\myexternaldocument}[1]{%
    \externaldocument{#1}%
    \addFileDependency{#1.tex}%
    \addFileDependency{#1.aux}%
}

\usepackage{xr}
\myexternaldocument{supplementary}

\usepackage{times}
\usepackage{latexsym}
\usepackage{subcaption}

\usepackage{amsmath}
\usepackage{array}
\usepackage{algorithm}
\usepackage{algorithmic}
\newcommand{\Q}{Q_i}

\newcommand{\tuple}[2]{(#1_1,#1_2,\dots,#1_{#2})}
\newcommand{\score}[2]{\lambda(#1, #2)}

\newcommand{\argminE}{\mathop{\mathrm{argmin}}}
\newcolumntype{L}{>{\centering\arraybackslash}m{5cm}}
\newcounter{fnote}
\newcommand{\fnote}[1] {
    \stepcounter{fnote}
    \footnotetext[\arabic{fnote}]{#1}
}

\usepackage[T1]{fontenc}
\usepackage[utf8]{inputenc}

\usepackage{microtype}

\title{'John ate 5 apples' != 'John ate some apples': Self-Supervised Paraphrase Quality Detection for Algebraic Word Problems}

\author{Rishabh Gupta\thanks{Equal Contribution} \and Venktesh V\footnotemark[1] \and
Mukesh Mohania \and
Vikram Goyal}
\tocauthor{Rishabh ~ Gupta, Venktesh ~ V,   Mukesh  ~ Mohania,  Vikram ~ Goyal}
\toctitle{Self-Supervised Paraphrase Quality Detection for Algebraic Word Problems}
\institute{Indraprastha Institute of Information Technology, Delhi \email{\{rishabh19089,venkteshv,mukesh,vikram\}@iiitd.ac.in}
}

\begin{document}
\maketitle
\begin{abstract}
This paper introduces the novel task of scoring paraphrases for Algebraic Word Problems (AWP) and presents a self-supervised method for doing so. In the current online pedagogical setting, paraphrasing these problems is helpful for academicians to generate multiple syntactically diverse questions for assessments. It also helps induce variation to ensure that the student has understood the problem instead of just memorizing it or using unfair means to solve it. The current state-of-the-art paraphrase generation models often cannot effectively paraphrase word problems, losing a critical piece of information (such as numbers or units) which renders the question unsolvable. There is a need for paraphrase scoring methods in the context of AWP to enable the training of good paraphrasers. Thus, we propose ParaQD, a self-supervised paraphrase quality detection method using novel data augmentations that can learn latent representations to separate a high-quality paraphrase of an algebraic question from a poor one by a wide margin. Through extensive experimentation, we demonstrate that our method outperforms existing state-of-the-art self-supervised methods by up to 32\% while also demonstrating impressive zero-shot performance.
\end{abstract}

\section{Introduction}
Algebraic Word Problems (AWPs) describe real-world tasks requiring learners to solve them using mathematical calculations. However, providing the same problem multiple times may result in the learner memorizing the mathematical formulation for the corresponding questions or exchanging the solution approach during exams without understanding the problem. Hence, paraphrasing would help prepare diverse questions and help to evaluate whether the student can arrive at the correct mathematical formulation and solution\footnotemark.

\begin{figure}[h]
    \centering
    \includegraphics[width=0.75\textwidth]{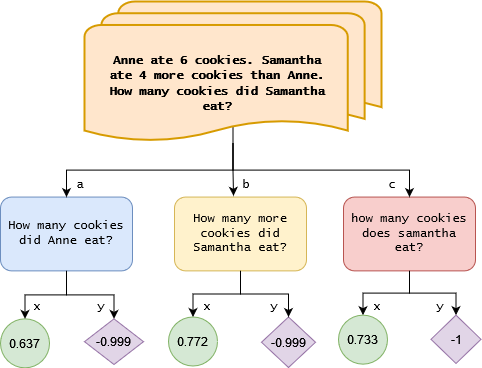}
    \caption[Caption]{Paraphrases by SOTA generation models. \textit{a} is output from PEGASUS fine-tuned on PAWS, \textit{b} is from T5 fine-tuned on Quora Question Pairs dataset and \textit{c} is from PARROT paraphraser built on T5. \textit{x} represents the cosine similarity scores assigned by the pretrained encoder MiniLM, while \textit{y} represents the scores with our proposed approach, ParaQD.}
    \label{fig:example}
\end{figure}
% \vspace{-0.4cm}
 
\fnote{\url{https://cutt.ly/MWqHsN8}}
% \fnote{Available at: \url{https://bit.ly/3F2c9vH}}
% \fnote{Available at: \url{https://bit.ly/3eVbAtk}}
% \fnote{Available at: \url{https://bit.ly/3qTLYCs}}
% \fnote{Available at: \url{https://bit.ly/3t0YS4u}}
 
The paraphrasing task can be tackled using supervised approaches like in \cite{egonmwan-chali-2019-transformer-seq2seq} or self-supervised approaches like in \cite{rotom}. As shown in Figure \ref{fig:example}, we observed that the generated paraphrases are of low quality as critical information is lost and the solution is not preserved. Some common issues that arose for the paraphrasing models were replacement or removal of numerical terms, important entities, replacement of units with irrelevant ones and other forms of information loss. These issues result in the generated question having a different solution or being rendered impossible to solve. Thus, there exists a need to automatically evaluate if a paraphrase preserves the semantics and solution of the original question. This is a \textit{more challenging problem} than detecting similarity for general sentences. The existing state-of-the-art semantic similarity models give a relatively high score even to very low-quality paraphrases of algebraic questions (where some critical information has been lost), as seen in Figure \ref{table:main} and Table \ref{table:main}. In Figure \ref{fig:example}, our approach ParaQD assigns the cosine similarity as -0.999, thereby preventing the low-quality paraphrases from getting chosen. There is a need for solutions like ParaQD because poor paraphrases of algebraic questions cannot be given to the students as they are either unsolvable (as observed in the figure) or do not preserve the original solution.
 
To tackle the issues mentioned above, we need a labelled dataset for training a proper scoring model. However, there does not exist a dataset for AWP with labelled paraphrases. Therefore, we propose multiple unsupervised data augmentations to generate positive and negative paraphrases for an input question. To model our negative augmentations, we identify crucial information in AWPs like numbers, units and key entities and design operators to perturb them. Similarly, for the positive augmentations, we design operators that promote diversity and retain the crucial information, thereby yielding a semantically equivalent AWP. On the other hand, existing augmentation methods like SSMBA \cite{ng2020ssmba} and UDA \cite{xie2020unsupervised} do not capture the crucial information in AWPs. Using the positive and negative paraphrases, we train a paraphrase scoring model using triplet loss. It explicitly allows for the separation of positives and negatives to learn representations that can effectively score paraphrases. In summary, our core contributions are :
 
 \begin{itemize}
 \setlength\itemsep{0.1em}
    \item We formulate a novel task of \textit{detecting paraphrase quality for AWPs}, which presents a different challenge than detecting paraphrases for general sentences.
     \item We propose a new unsupervised data augmentation method that drives our paraphrase scoring model, \textit{ParaQD}.
     \item We demonstrate that our method leads to a scoring model that surpasses the existing state-of-the-art text augmentation methods like SSMBA and UDA.
     \item We evaluate ParaQD using test sets prepared using operators disjoint from train augmentation operators and observe that ParaQD demonstrates good performance. We also demonstrate the zero-shot performance of ParaQD on new AWP datasets.
 \end{itemize}
Code and Data are available at: \url{https://github.com/ADS-AI/ParaQD}
% \fnote{Code and Data are available at: \url{https://anonymous.4open.science/r/ParaQD-81B0/}}

\section{Related Work}
This section briefly discusses prior work in text data augmentation methods. One of the notable initial works in data augmentation for text \cite{zhang2016characterlevel} replaced words and phrases with synonyms to obtain more samples for text classification. In the work \cite{xie2017data}, the authors propose noising methods for augmentations where words are replaced with alternate words based on unigram distribution, but it introduces a noising parameter. A much easier text augmentation method, EDA, was proposed in the work \cite{wei2019eda}. The authors propose several operators such as random word deletion and synonym replacement to generate new sentences. The above works are based on heuristics and depend on a hyperparameter for high-quality augmentations. 

More recently, self-supervised text augmentation methods have provided a superior performance on multiple tasks. In UDA \cite{xie2020unsupervised}, the authors propose two text augmentation operators, namely backtranslation and TF-IDF based word replacement, where words with low TF-IDF scores are replaced. In SSMBA \cite{ng2020ssmba}, the authors propose a manifold-based data augmentation method where the input sentences are projected out of the manifold by corrupting them with token masking, followed by a reconstruction function to project them back to the manifold. Another self-supervised augmentation method named InvDA (Inverse Data Augmentation) was proposed in Rotom \cite{rotom} which was similar to SSMBA in that it tried to reconstruct the original sentence from the corrupted version. Several rule-based text augmentation methods have also been proposed, like \cite{kang2018adventure} which uses Natural Language Inference (NLI) for augmentation, and \cite{asai2020logicguided} leverages linguistic knowledge for the question-answering task. 
% Applying these methods directly to algebraic questions is not as beneficial because they may result in loss of crucial information rendering the question unsolvable when generating a positive paraphrase.

\section{Methodology}

In this section, we describe the proposed method for paraphrase quality detection for algebraic word problems. The section is divided into two components: Data Augmentation and Paraphrase Quality Detection.
% This section consists of two subsections:
% \begin{itemize}
%     \item\textbf{Data Augmentation:} Here, we detail the data augmentation techniques used for generating positive and negative samples.
%     \item\textbf{Paraphrase Quality Detection:} We detail our approach for detecting the quality of question paraphrases.
% \end{itemize}

\subsection{Data Augmentation}

% Define positive and negative paraphrases here. Maybe use the example of the pretrained models (Pegasus, T5) to illustrate the point.

For data augmentation, we define 10 distinct operators to generate the training set. Out of the 10, 4 are positive (i.e. information preserving) transformations, and 6 are negative (information perturbing) transformations. Our negative operators are carefully chosen after observing the common mistakes made by various paraphrasing models to \textit{explicitly teach} the quality detection model to assign a low score for incorrect paraphrases.

Let $Q = \{Q_1, Q_2, Q_3, \dots Q_n\}$ denote the set of questions. Each question $Qi$ can be tokenized into sentences $Q_{i1}, Q_{i2} \dots Q_{ip}$ where $p$ denotes the number of sentences in question $Q_i$. Let an augmentation be denoted by a function $f$, such that $f_{i}(Q_{j})$ represents the output of the $i$th augmentation on the $j$th question.

% representation of Q_i'?
The function $\lambda: Q \times Q \mapsto \{0, 1\}$ represents a labelling function which returns 1 if the input ($Q_i$, $Q_i'$) is a valid paraphrase, and 0 if not. Based on the design of our augmentations (explained in the next section), we work under the following assumption for the function $f$:
  \begin{align*}
    \lambda(Q_a, f_i(Q_a))=
    \begin{cases}
      1,&   1\leq i \leq4\\
      0, &  5\leq i \leq10
    \end{cases}
  \end{align*}
  
For the purposes of explanation, we will use a running example with question $\mathbf{Q_0} = $ \textit{Alex travelled 100 km from New York at a constant speed of 20 kmph. How many hours did it take him in total?} \\
% In the next section, we will detail both the positive and negative augmentation techniques that we used. 
\vspace{-0.3cm}
\subsection{Positive Augmentations}

\subsubsection{$f_1$: Backtranslation}
\label{f1}

Backtranslation is the procedure of translating an example $Q_i$ from language $A$ to language $B$, and then translating it back to language $A$, yielding a paraphrase $Q_i'$. In our case, given an English question $Q_i$ comprised of precisely $p$ sentences $Q_{i1} \dots Q_{ip}$, we translate each sentence $Q_{ij}$ to German $Q_{ij}^*$, and then translate $Q_{ij}^*$ back to English yielding $Q_{ij}'$ $\forall j \in \{1, 2,\dots p\}$. Further details are provided in Appendix \ref{appendix:augs}.
\begin{align*}
f_1({Q_i}) = concat(Q_{i1}', Q_{i2}' \dots Q_{ip}')    
\end{align*}
$\mathbf{f_{1}(Q_0)}$ :  \textit{Alex was driving 100 km from New York at a constant speed of 20 km / h. How many hours did it take in total?}

\subsubsection{$f_2$: Same Sentence}
\label{f2}
Inspired by SimCSE \cite{gao2021simcse}, we explicitly provide the same sentence as a positive augmentation as the standard dropout masks in the encoder act as a form of augmentation. \\
$\mathbf{f_{2}(Q_0)}$ :  \textit{Alex travelled 100 km from New York at a constant speed of 20 kmph. How many hours did it take him in total?}

\subsubsection{$f_3$: Num2Words}
\label{f3}
Let $\alpha$ be a function that converts any number to its word form. Given a question $Q_i$, we extract all the numbers $N_i = \{n_{i1}, n_{i2} \dots n_{ik}\}$ from $Q_i$. For each number $n_{ij} \in N_i$, we generate its word representation $\alpha(n_{ij})$, and replace $n_{ij}$ by $\alpha(n_{ij})$ in $Q_i$ to get $f_3(\Q)$. This is done because paraphrasing models can replace numbers with their word form, and thus to ensure the scoring model does not consider it as a negative, we explicitly steer it to consider it a positive. \\
$\mathbf{f_{3}(Q_0)}$:  \textit{Alex travelled one hundred km from New York at a constant speed of twenty kmph. How many hours did it take him in total?}

\subsubsection{$f_4$: UnitExpansion}
\label{f4}
Let $\upsilon$ be a function that converts the abbreviation of a unit into its full form. We detect all the abbreviated units $U_i = \{u_{i1}, u_{i2} \dots u_{ik}\}$ from $Q_i$ (using a predefined vocabulary of units and regular expressions). For each unit $u_{ij} \in U_i$, we generate its expansion $\upsilon(u_{ij})$, and replace $u_{ij}$ by $\upsilon(u_{ij})$ in $Q_i$. This transformation helps the model to learn the units and their expansions, and consider them as the same when scoring a paraphrase. \\
$\mathbf{f_{4}(Q_0)}$:  \textit{Alex travelled 100 kilometre from New York at a constant speed of 20 kilometre per hour. How many hours did it take him in total?}

\subsection{Negative Augmentations}

\subsubsection{$f_5$: Most Important Phrase Deletion}
\label{f5}
The removal of unimportant words like stopwords (the, of, and) from an algebraic question will not perturb the solution or render it impossible to solve. 

Thus, to generate hard negatives, we chose the most critical phrase, $p_{imp}$ in any question, deleting which would generate $Q_i'$ such that $\score{\Q}{Q_i'} = 0$. Let $\Psi: Q \mapsto P$ denote a function which returns the set of $k$ most critical phrases $\tuple{p}{k}$ in the input $\Q$. 
\begin{align*}
    &p_{imp} = \argminE_p(cossim(\Q, \Q\backslash p)) & \forall p \in \Psi(\Q) \\
    &f_5(\Q) = \Q\backslash p_{imp}
\end{align*}
where $cossim$ denotes cosine similarity and $\Q\backslash p$ denotes the deletion of $p$ from $\Q$. Further details are present in Appendix \ref{appendix:augs}.\\
$\mathbf{f_{5}(Q_0)}$:  \textit{Alex travelled 100 km from New York at a constant speed of 20 kmph. How did it take him in total?}

\subsubsection{$f_6$: Last Sentence Deletion}
\label{f6}
When using existing paraphrasing models such as Pegasus, the last few words or even the complete last sentence of the input question got deleted in the generated paraphrase in some cases. Thus, to account for this behaviour, we use this transformation as a negative. More formally, let the input $\Q$ be tokenized into $p$ sentences $Q_{i1}, Q_{i2} \dots Q_{ip}$ and the sentence $Q_{i1}$ be tokenized into k tokens $Q_{i11}, Q_{i12} \dots Q_{i1k}$. Then,
  \begin{align*}
    &f_6(\Q)=
    \begin{cases}
      concat(Q_{i11}, Q_{i12} \dots Q_{i1(k-3)}) & p=1 \\
      concat(Q_{i1}, Q_{i2} \dots Q_{i(p-1)}) & p>1 
     \end{cases}
  \end{align*} \\
$\mathbf{f_{6}(Q_0)}$:  \textit{Alex travelled 100 km from New York at a constant speed of 20 kmph.}

\subsubsection{$f_7$: Named Entity Replacement}
\label{f7}
Since named entities are an important part of questions, we either replace them with a random one of the same category (from a precompiled list) or with the empty string (deletion). Let $\epsilon: Q \mapsto E$ denote a function which returns a set of all named entities present in the input $\Q$, such that $\tuple{e}{k} = \epsilon(\Q)$. We randomly sample $w$ elements $E_i = (e_a, e_b \dots e_w)$ from $\tuple{e}{k}$ and replace/delete the entities. We set $w = rand(1, min(3, k))$ where $rand(a,b)$ represents the random selection of a number from $a$ to $b$ (inclusive). This restricts $w$ from being more than 3, thus increasing the difficulty of the generated negative. \\
$\mathbf{f_{7}(Q_0)}$:  \textit{Sarah travelled 100 km from at a constant speed of 20 kmph. How many hours did it take him in total?}

\subsubsection{$f_8$: Numerical Entity Deletion}
\label{f8}
Since numbers are critical to algebraic questions, their removal perturbs the solution and helps generate hard negatives. Let $\nu: Q \mapsto N$ represent a function which returns a set of all numbers present in the input $\Q$, such that $\tuple{n}{k} = \nu(\Q)$. We randomly sample a subset of numbers $N_i$ from $\tuple{n}{k}$, and sample a string $s$ from $S$ = \textit{("some", "a few", "many", "a lot of", "")}. For each number $n_{j} \in N_i$, we replace it by $s$ in $Q_i$. We set $|max(N_i)| = 2$. Similar to $f_7$, this makes it more challenging for the scoring model as we don't necessarily delete all the numbers, thereby generating harder negatives. This allows the model to learn that even the loss of one number renders the resultant output as an invalid paraphrase, thus getting assigned a low score.\\
$\mathbf{f_{8}(Q_0)}$:  \textit{Alex travelled some km from New York at a constant speed of some kmph. How many hours did it take him in total?}

\subsubsection{$f_{9}$: Pegasus}
\label{f9}
Pegasus \cite{zhang2019pegasus} is a transformer-based language model, fine-tuned on PAWS \cite{zhang2019paws} for our purpose. Pegasus consistently gave poor results for paraphrasing algebraic questions, as shown in Figure \ref{fig:example}. This provided the impetus for using it to generate hard negatives. \\
$\mathbf{f_{9}(Q_0)}$: = \textit{The journey from New York to New Jersey took Alex 100 km at a constant speed.}

\subsubsection{$f_{10}$: UnitReplacement}
\label{f10}
Paraphrasing models sometimes have a tendency to replace units with similar ones (such as \textit{feet} to \textit{inches}). Since this would change the solution to an algebraic question, we defined this transformation to replace a unit with a different one from the same category. We identified 5 categories, $C$ = \textit{[Currency, Length, Time, Weight, Speed]} to which most units appearing in algebraic problems belong. Our transformation was defined such that a unit $u_a$ belonging to a particular category $C_i$ is replaced with a unit $u_b$, such that $u_b \in C_i$ and $u_a \neq u_b$. For instance, \textit{hours} could get converted to \textit{minutes} or \textit{days}, \textit{grams} could get converted to \textit{kilograms}.

Let $C$ be the set of identified unit categories and $\Upsilon: U \mapsto U$ be a function that takes as input unit $u_a \in C_i$ and returns a different unit $u_b \in C_i$,  where $C_i \in C$. Given the input $\Q$ containing units $U_i = (u_a, u_b \dots u_n)$, we sample a set of units $U_{is} = \{u_x, \dots u_z\}$ and replace them with $\{\Upsilon(u_i)$ $\forall u_i \in U_{is}\}$ to generate $f_{10}(\Q)$. \\
$\mathbf{f_{10}(Q_0)}$: \textit{Alex travelled 100 m from New York at a constant speed of 20 kmph. How many hours did it take him in total?}

In the next section, we will detail our approach to training a model to detect the quality of paraphrases and how it can be used to score paraphrases.

\subsection{Paraphrase Quality Detection}
For detecting the quality of the paraphrases, we use MiniLM \cite{wang2020minilm} as our base encoder (specifically, the version with 12 layers which maps the input sentences into 384-dimensional vectors)\footnotemark. We utilize the implementation from SentenceTransformers \cite{reimers-gurevych-2019-sentence}, where the encoder was trained for semantic similarity tasks using over a billion training pairs and achieved high performance with a fast encoding speed\footnotemark.

\fnote{\url{https://bit.ly/3F2c9vH}}
\fnote{\url{https://sbert.net/docs/pretrained\_models.html}}

We train the model using triplet loss. For each question $\Q$, let the positive transformation $\Q^+$ be denoted by $pos(\Q)$ and the negative transformation $\Q^-$ by $neg(\Q)$ where $pos \in (f_1, \dots f_4)$ and $neg \in (f_5, f_6 \dots f_{10})$. Let the vector representation of any question $\Q$ when passed through the encoder be denoted as $ENC(Q)$. Then the loss is defined as
\begin{align*}
    Loss(Q, Q^+, Q^-) = \sum_{i} max(0, \alpha - dist(\Q, \Q^-) + dist(\Q, \Q^+))
\end{align*}
where $\alpha$ is the margin parameter, $dist(\Q,\Q^l) = 1 - cossim(ENC(\Q), ENC(\Q^l))$ and $l \in \{+,-\}$. The loss ensures that the model yields vector representations such that the distance between $Q_i$ and $Q_i^+$ is smaller than the distance between $Q_i$ and $Q_i^-$.

At inference time, to obtain the paraphrase score of $\Q$ and $\Q'$, we use cosine similarity. Let $score : Q \times Q \mapsto [-1, 1]$ denote the scoring function, then for a pair of questions $(\Q, \Q')$:
\begin{align*}
    &\rho_i, \zeta_i = ENC(\Q), ENC(\Q') \\
    &score(\Q, \Q') = cossim(\rho_i, \zeta_i) = \frac{\rho_i\cdot\zeta_i}{|\rho_i|\cdot|\zeta_i|}
\end{align*}

\section{Experiments}

All the experiments were performed using a Tesla T4 and P100. All models, including the baselines, were trained for 9 epochs with a learning rate of 2e-5 using AdamW as the optimizer with seed 3407. We used a linear scheduler, with 10\% of the total steps as warm-up having a weight decay of 0.01.

\subsection{Datasets}
The datasets used in the experiments are:

\textbf{AquaRAT} \cite{ling2017program} (Apache, V2.0) is an algebraic dataset consisting of 30,000 (post-filtering) problems in the training set, 254 problems for validation and 220 problems for testing. After applying the test set operators to yield paraphrases, we get 440 samples for testing with manual labels.

\textbf{EM\_Math} is a dataset consisting of mathematics questions for students from grades 6-10 from our partner company ExtraMarks. There are 10,000 questions in the training set and 300 in the test set. After applying the test operators, we get 600 paraphrase pairs. 

\textbf{SAWP} (Simple Arithmetic Word Problems) is a dataset that we collected (from the internet) consisting of 200 algebraic problems. We evaluate the proposed methods in a zero-shot setting on this dataset by using the model trained on the AquaRAT dataset. After applying the test set operators, we get 400 paraphrase pairs.
\vspace{-0.02cm}

\textbf{PAWP} (Paraphrased Algebraic Word Problems) is a dataset of 400 algebraic word problems collected by us. We requested two academicians from the partnering company (paid fair wages by the company) to manually write paraphrases (both valid and invalid) rather than using our test set operators. We use this dataset for zero-shot evaluation to demonstrate the performance of our model on human-crafted paraphrases. \\
Our data can also be used as a \textbf{seed set} for the task of paraphrase generation for algebraic questions.

\subsection{Test Set Generation}
For generating the synthetic test set (for AquaRAT, EM\_Math and SAWP), we define a different set of operators to generate positive and negative paraphrases to test the ability of our method to generalize to a different data distribution. For any question $\Q$ in the test set, we generate two paraphrases and manually annotate the question-paraphrase pairs with the help of two annotators. The annotators were instructed to mark valid paraphrases as 1 and the rest as 0. We observed Cohen's Kappa values of \textbf{0.79}, \textbf{0.84} and \textbf{0.70} on AquaRAT, EM\_Math and SAWP, respectively, indicating a substantial level of agreement between the annotators.

\subsubsection{Operator Details}
We defined two positive ($f_a$, $f_b$) and three negative ($f_c$, $f_d$, $f_e$) test operators. For each question, we randomly chose one operator from each category for generating paraphrases. These functions are:

\textbf{$f_a$: Active-Passive}: We noticed that most algebraic questions are written in the active voice. We used a transformer model for converting them to passive voice\footnotemark, followed by a grammar correction model\footnotemark on top of this to ensure grammatical correctness.

\fnote{\url{https://bit.ly/3FbPIEu}}
\fnote{\url{https://bit.ly/3HGOMcQ}}

\textbf{$f_b$:  Corrupted Sentence Reconstruction}: We corrupt an input question by shuffling, deleting and replacing tokens, similar to ROTOM \cite{rotom} but with additional leniency (Appendix \ref{appendix:augs}). We then train a sequence transformation model (t5-base) to reconstruct the original question from the corrupted one, which yields a paraphrase.

\textbf{$f_c$:  TF-IDF Replacement}: Instead of the usual replacement of words with low TF-IDF score \cite{xie2020unsupervised},  we replace the words with high TF-IDF scores with random words in the vocabulary. This helps us generate negative paraphrases as it removes the meaningful words in the original question rendering it unsolvable.  

\textbf{$f_d$: Random Deletion}:  Random deletion is the process of randomly removing some tokens in the input example \cite{wei2019eda} to generate a paraphrase. 

\textbf{$f_e$: T5}: We used T5 \cite{raffel2020exploring} fine-tuned on Quora Question Pairs to generate negatives as it was consistently resulting in paraphrases with missing information (Figure \ref{fig:example}).

\subsection{Baselines}
% We compare our method against two other data augmentation methods: UDA \cite{xie2020unsupervised}, and SSMBA \cite{ng2020ssmba}. We also compare it against the pre-trained MiniLM \cite{wang2020minilm} model to see the effect of our data augmentation in scoring the paraphrase quality of algebraic questions. To measure the strength of the data augmentation, we keep the encoder constant throughout all baselines. Instead of using the baselines as is, we enhance them by using  MPNet to soft label the augmented samples. More formally, given an input $\Q$ and a paraphrase $\Q'$, we use MPNet to determine whether $\Q'$ is a positive or negative paraphrase of $\Q$ as follows:
%   \begin{align*}
%     &\alpha_i, \beta_i = MPNet(\Q), MPNet(\Q') \\
%     &\lambda(\Q, \Q')=
%     \begin{cases}
%       1 &if\;cossim(\alpha_i, \beta_i) > \tau \\
%       0 &if\;cossim(\alpha_i, \beta_i) \leq \tau 
%      \end{cases}
%   \end{align*}
% where $\tau$ is the threshold for the cosine similarity. For our experiments, we found that $\tau = 0.8$ worked well.
We compare against two SOTA data augmentation methods, UDA and SSMBA. For all the baselines, we use the same encoder (MiniLM) as for our method to maintain consistency across the experiments and enable a fair comparison.

\textbf{UDA}: UDA uses backtranslation and TF-IDF replacement (replacing words having a low score) to generate augmentations for any given input.

\textbf{SSMBA}: SSMBA is a data augmentation technique that uses corruption and reconstruction functions to generate the augmented output. The corruption is performed by masking some tokens in the input and using an encoder (such as BERT \cite{BERT}) to fill the masked token. 

Since the baselines are intended to generate positive paraphrases, we consider other questions in the dataset (in-batch) as negatives to train using the triplet loss. Alongside the direct implementation of UDA and SSMBA, we also compare pseudo-labelled versions of these baselines. The version of baselines without pseudo-labelling is used in all the experiments unless stated with suffix \textit{(with pl)}. The details of pseudo labelling are provided in Appendix \ref{appendix:pl}.

% \vspace{-10mm}

\subsection{Metrics}
\label{metrics}
% \vspace{-2em}
Our main goal is to ensure the separation of valid and invalid paraphrases by a wide margin. This allows for extrapolation to unseen and unlabelled data (the distribution of scores for positive and negative paraphrases is unknown, thus threshold can be set to the standard 0.5 or a nearby value due to wider margins). It allows for the score to be used as a selection metric using maximization strategies like Simulated Annealing \cite{liu-etal-2020-unsupervised} or as reward using Reinforcement Learning \cite{stiennon2020learning,yasui-etal-2019-using} to steer generation. To this end, along with Precision, Recall, and F1 (both macro and weighted), we compute the separation between the mean positive and mean negative scores. More formally, let the score of all $(\Q, \Q^+)$ pairs be denoted by $score(Q, Q^+)$ and the score of all $(\Q, \Q^-)$ pairs be denoted by $score(Q, Q^-)$ where $\lambda(\Q, \Q^+) = 1$ and $\lambda(\Q, \Q^-) = 0$. Then,
% \vspace{-0.7cm}
\begin{align*}
    &\mu^s\; (separation) = \mu^+ - \mu^- \\
    &\mu^l = E[score(Q, Q^l)]\; \forall\; l \in \{+,-\}
\end{align*}
% \vspace{-2.4cm}

\subsection{Test Set Details} \label{test}
\vspace{-0.3em}
The number of positive and negative pairs are (139, 301) in AquaRAT, (223, 377) in EM, (130, 270) in SAWP and (199, 201) in PAWP. The details of the success of test set operators are shown in the form of confusion matrices in Figure \ref{fig:test} (supplementary). The average precision, recall and accuracy of the operators across the datasets are 0.4, 0.59 and 0.56. The low precision is due to the inability of positive operators to generate valid paraphrases consistently, as the task of effectively paraphrasing algebraic questions is challenging. This further demonstrates the usefulness of a method like ParaQD that can be effectively used to distinguish the paraphrases as an objective to guide paraphrasing models (\ref{metrics}).

\begin{table*}[h!]
% \small
\caption{Precision, Recall, F1 and Separation across all methods and datasets.}
\begin{tabular}{clrrrrrrrrp{0.6cm}}
% \small
\hline
\multicolumn{1}{l}{\multirow{2}{*}{\textit{\textbf{Dataset}}}} & \multirow{2}{*}{\textbf{Method}} & \multicolumn{3}{c}{\textbf{Macro}}                                     & \multicolumn{3}{c}{\textbf{Weighted}}                                           & \multicolumn{1}{l}{\multirow{2}{*}{\textbf{$\mu^+$}}} & \multicolumn{1}{l}{\multirow{2}{*}{\textbf{$\mu^-$}}} & \multicolumn{1}{l}{\multirow{2}{*}{\textbf{$\mu^s$}}} \\ \cline{3-8}
\multicolumn{1}{l}{}                                           &                                  & \multicolumn{1}{l}{P} & \multicolumn{1}{l}{R} & \multicolumn{1}{l}{F1} & \multicolumn{1}{l}{P} & \multicolumn{1}{l}{R} & \multicolumn{1}{l}{F1} & \multicolumn{1}{l}{}                          & \multicolumn{1}{l}{}                          & \multicolumn{1}{l}{}                            \\ \hline
\multirow{6}{*}{AquaRAT}                                       & Pretrained                       & 0.658                 & 0.502                 & 0.569                  & 0.784                 & 0.318                 & 0.453                  & 0.977                                         & 0.897                                         & 0.080                                            \\ 
                                                              & UDA                     & 0.661                 & 0.512                 & 0.577                  & 0.786                 & 0.332                 & 0.467                  & 0.995                                         & 0.966                                         & 0.029                                           \\  
                                                              & UDA (w pl)                       & 0.659                 & 0.507                 & 0.573                  & 0.785                 & 0.325                 & 0.460                   & 0.996                                         & 0.973                                         & 0.023                                           \\  
                                                              & SSMBA                            & 0.645                 & 0.554                 & 0.596                  & 0.757                 & 0.395                 & 0.520                   & 0.965                                         & 0.829                                         & 0.137                                           \\ 
                                                              & SSMBA (w pl)                     & 0.663                 & 0.522                 & 0.584                  & 0.787                 & 0.345                 & 0.480                   & 0.997                                         & 0.928                                         & 0.069                                           \\ 
                                                              & ParaQD (ours)                    & 0.678                 & 0.695                 & \textbf{0.687}         & 0.762                 & 0.625        & \textbf{0.687}                  & 0.770                                          & -0.010                                         & \textbf{0.780}                                   \\ \hline
\multirow{6}{*}{EM\_Math}                                      & Pretrained                       & 0.694                 & 0.534                 & 0.604                  & 0.773                 & 0.415                 & 0.540                   & 0.955                                         & 0.796                                         & 0.158                                           \\ 
                                                              & UDA                     & 0.648                 & 0.523                 & 0.579                  & 0.716                 & 0.403                 & 0.516                  & 0.991                                         & 0.912                                         & 0.079                                           \\  
                                                              & UDA (w pl)                       & 0.683                 & 0.587                 & 0.631                  & 0.751                 & 0.485                 & 0.589                  & 0.963                                         & 0.751                                         & 0.213                                           \\ 
                                                              & SSMBA                            & 0.615                 & 0.564                 & 0.588                  & 0.669                 & 0.470                  & 0.552                  & 0.871                                         & 0.729                                         & 0.142                                           \\  
                                                              & SSMBA (w pl)                     & 0.655                 & 0.586                 & 0.619                  & 0.716                 & 0.492                 & 0.583                  & 0.937                                         & 0.629                                         & 0.308                                           \\  
                                                              & ParaQD (ours)                    & 0.665                 & 0.665                 & \textbf{0.665}         & 0.708                 & 0.622                 & \textbf{0.662}         & 0.667                                         & 0.012                                         & \textbf{0.655}                                  \\ \hline
\multirow{6}{*}{SAWP}                                          & Pretrained                       & 0.162                 & 0.500                   & 0.245                  & 0.106                 & 0.325                 & 0.159                  & 0.964                                         & 0.896                                         & 0.068                                           \\ 
                                                              & UDA                      & 0.557                 & 0.514                 & 0.535                  & 0.636                 & 0.358                 & 0.458                  & 0.958                                         & 0.912                                         & 0.046                                           \\  
                                                              & UDA (w pl)                       & 0.667                 & 0.519                 & 0.583                  & 0.783                 & 0.350                  & 0.484                  & 0.990                                          & 0.929                                         & 0.061                                           \\  
                                                              & SSMBA                            & 0.662                 & 0.594                 & 0.626                  & 0.763                 & 0.460                  & 0.574                  & 0.929                                         & 0.758                                         & 0.172                                           \\  
                                                              & SSMBA (w pl)                     & 0.649                 & 0.537                 & 0.588                  & 0.757                 & 0.378                 & 0.504                  & 0.978                                         & 0.864                                         & 0.115                                           \\  
                                                              & ParaQD (ours)                    & 0.636                 & 0.645                 & \textbf{0.640}          & 0.709                 & 0.582                 & \textbf{0.640}          & 0.656                                         & 0.068                                         & \textbf{0.589}                                  \\ \hline
\multirow{6}{*}{PAWP}                                          & Pretrained                       & 0.749                 & 0.502                 & 0.602                  & 0.751                 & 0.500                   & 0.600                    & 0.948                                         & 0.905                                         & 0.042                                           \\  
                                                              & UDA                     & 0.558                 & 0.507                 & 0.532                  & 0.559                 & 0.505                 & 0.530                   & 0.960                                          & 0.948                                         & 0.012                                           \\  
                                                              & UDA (w pl)                       & 0.668                 & 0.510                  & 0.578                  & 0.669                 & 0.507                 & 0.577                  & 0.988                                         & 0.961                                         & 0.026                                           \\ 
                                                              & SSMBA                            & 0.536                 & 0.512                 & 0.524                  & 0.536                 & 0.510                  & 0.523                  & 0.874                                         & 0.853                                         & 0.021                                           \\  
                                                              & SSMBA (w pl)                     & 0.551                 & 0.510                  & 0.530                   & 0.552                 & 0.507                 & 0.529                  & 0.939                                         & 0.913                                         & 0.026                                           \\  
                                                              & ParaQD (ours)                    & 0.703                 & 0.669                 & \textbf{0.685}         & 0.703                 & 0.668                 & \textbf{0.685}         & 0.749                                         & 0.076                                         & \textbf{0.673}                                  \\ \hline
\end{tabular}
% \vspace{-4mm}

\label{table:main}
\end{table*}

\begin{table}[]
\caption{Summarizing the top-2 positive (Op+) and negative (Op-) operators across datasets. }
\begin{tabularx}{\columnwidth}{cXXXX}
\multirow{2}{*}{\textbf{Dataset}} & \multicolumn{2}{c}{\textbf{Op+}} & \multicolumn{2}{c}{\textbf{Op-}} \\ \cline{2-5} 
         & \textbf{1} & \textbf{2} & \textbf{1} & \textbf{2} \\ \hline
AquaRAT  & $f_3$         & $f_1$         & $f_9$         & $f_5$         \\
EM\_Math & $f_4$         & $f_1$         & $f_9$         & $f_8 $        \\
SAWP     & $f_2$         & $f_1$         & $f_9 $        & $f_6$         \\
PAWP     & $f_1 $        & $f_2$         & $f_{10}$        & $f_9$         \\ \hline
\end{tabularx}

% Here, Op+ and Op- refer to positive and negative operators respectively.
\label{table:top2}
\end{table}

\section{Results and Analysis}
The performance comparison and results of all methods are shown in Table \ref{table:main}. Across all datasets, for the measures macro-F1, weighted-F1 and separation, ParaQD outperforms all the baselines by a significant margin. For instance, the margin of separation in ParaQD is 5.69 times the best baseline SSMBA. To calculate the precision, recall and F1 measures, we threshold the obtained scores at the standard $\tau = 0.5$. Since this is a self-supervised method, there are no human-annotated labels available for the training and validation set. This means that the distribution of scores is unknown, and thus, the threshold can not be tuned on the validation set.

\subsection{Performance}
Our primary metric is separation (for reasons detailed in \ref{metrics}). Weighted F1 is more representative of the actual performance than macro F1 due to imbalanced data (\ref{test}), and the results are discussed further.
\subsubsection{AquaRAT and EM\_Math}:
ParaQD outperforms the best-performing baseline by 32.1\% weighted F1 on AquaRAT and 12.4\% weighted F1 on EM\_Math. The separation achieved by ParaQD on AquaRAT is 0.78 while the best performing baseline achieves 0.137, and on EM\_Math, our method achieves a separation of 0.655 while the best performing baseline achieves a separation of 0.308.
\\
\textbf{SAWP}:
Evaluating zero-shot performance on SAWP, ParaQD outperforms the best performing baseline by 11.5\% weighted F1 and achieves a separation of 0.589 as compared to the 0.172 achieved by the best baseline. This demonstrates the ability of our method to perform well even on zero-shot settings, as the distribution of this dataset is not identical to the ones that the model was trained on.\\
\textbf{PAWP}:
Our method beats the best performing baseline by 14\% weighted F1 on the manually created dataset PAWP, which also consists of a zero-shot setting. It demonstrates an impressive separation of 0.673, while the best performing baseline only has a separation of 0.042. This is practically applicable as it highlights that our method can also be used to evaluate paraphrases that have been manually curated by academicians (especially on online learning platforms) instead of only on automatically generated paraphrases.

To analyze and gain a deeper insight into these results, we plotted the confusion matrices (Figure \ref{fig:cm}), and observed that ParaQD is able to consistently recognize invalid paraphrases to a greater extent than the baselines as it learns to \textit{estimate the true distribution of negative samples} more effectively through our novel data augmentations.

\begin{figure*}[hbt!]
% \centering
% \hspace{-2pt}
  \begin{subfigure}[b]{\textwidth}
        \centering
        % \hspace{-45pt}
        \includegraphics[width=0.55\textwidth]{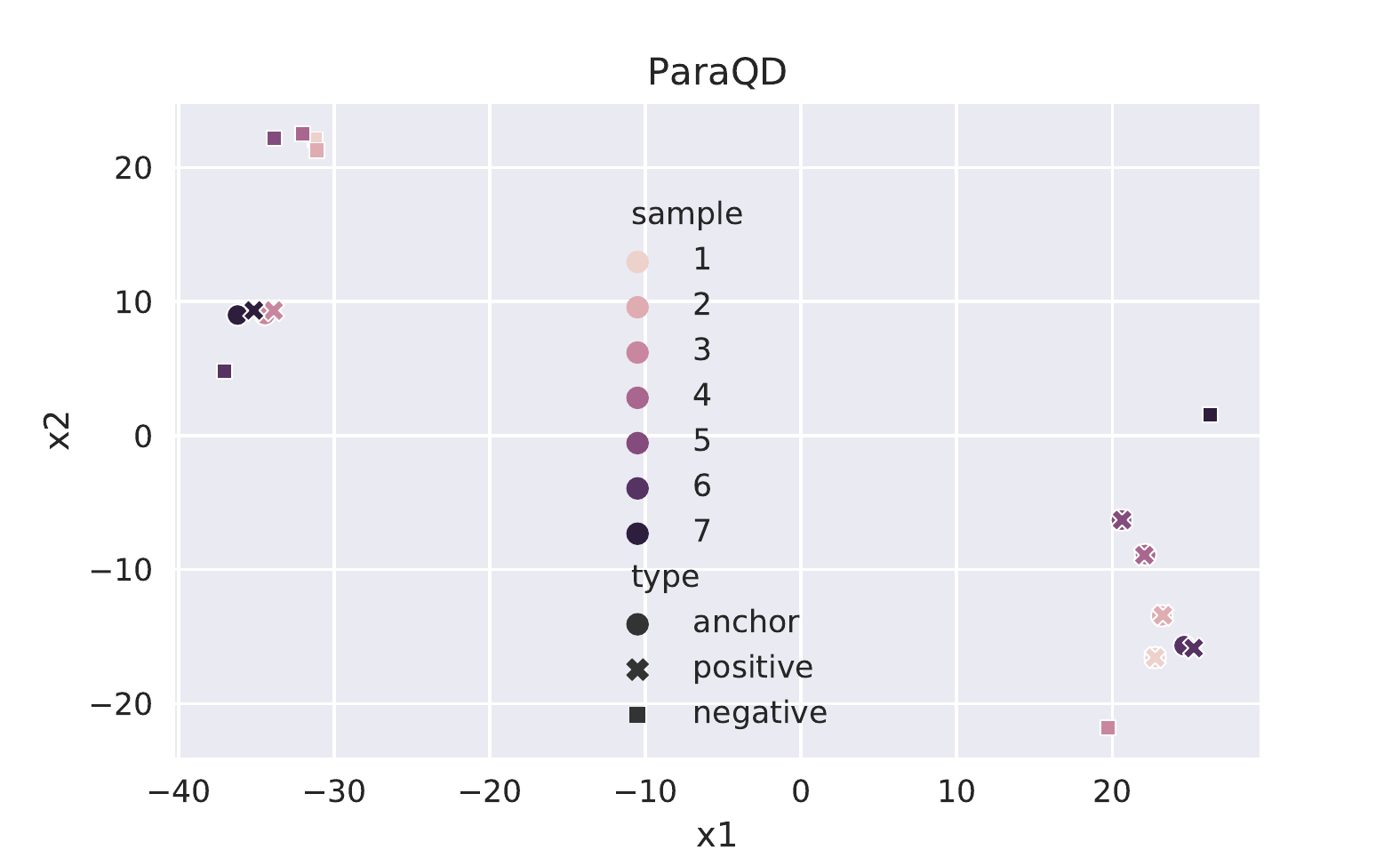}%
        \includegraphics[width=0.55\textwidth]{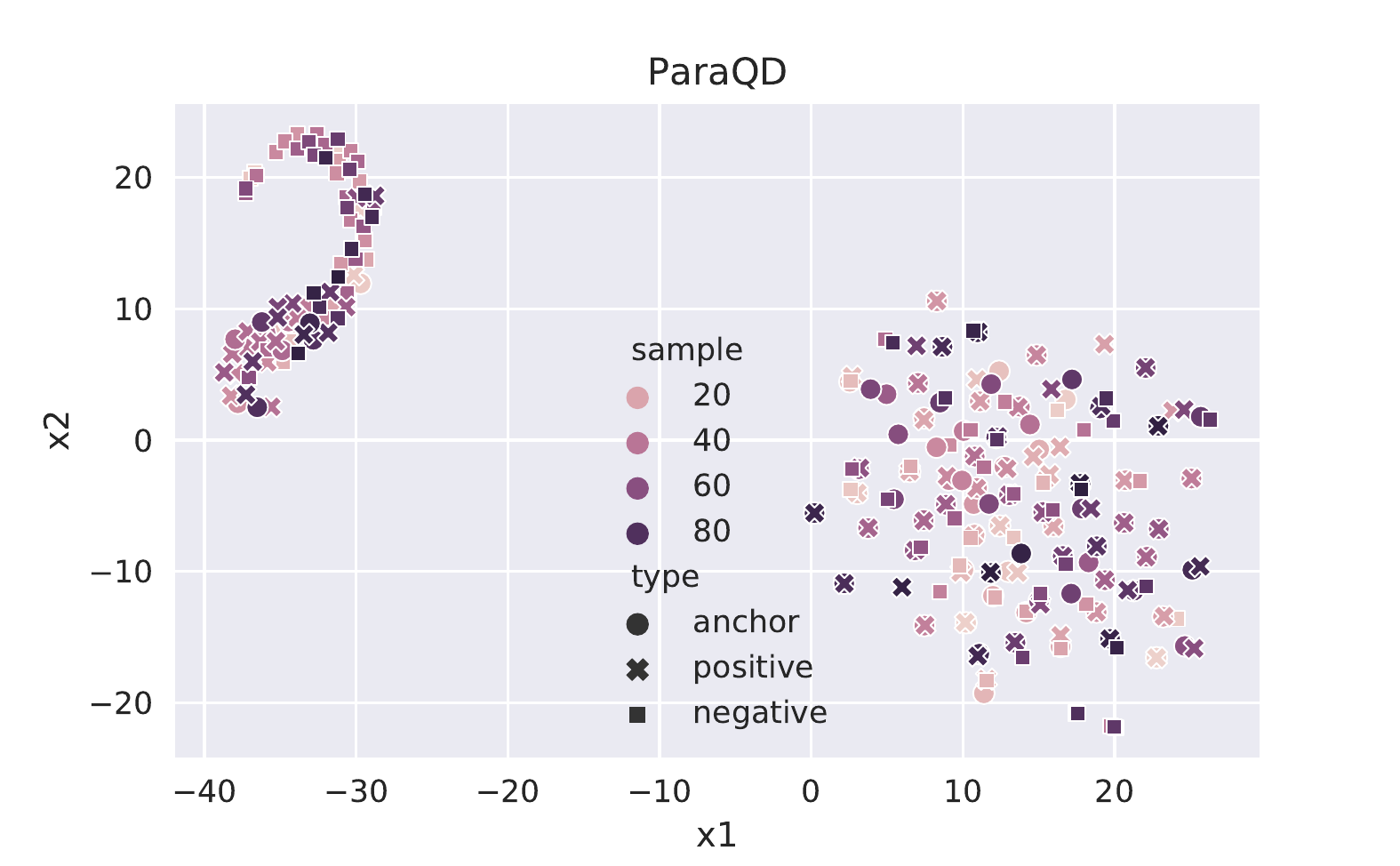}
        \caption{ParaQD}
        \label{fig:plots-para}
    \end{subfigure}
  \begin{subfigure}[b]{\textwidth}
        \centering
        % \hspace{-45pt}
        \includegraphics[width=0.55\textwidth]{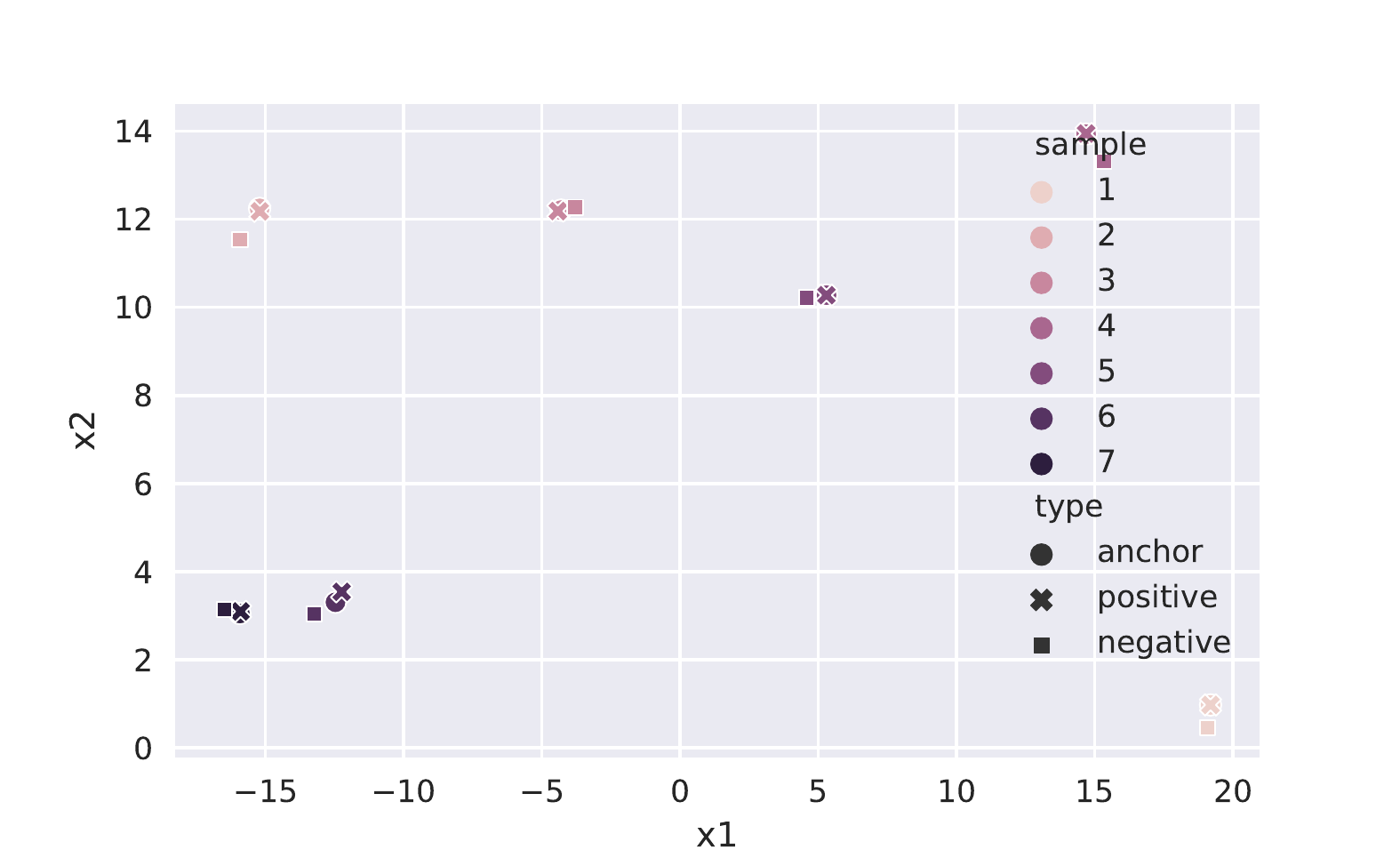}%
        \includegraphics[width=0.55\textwidth]{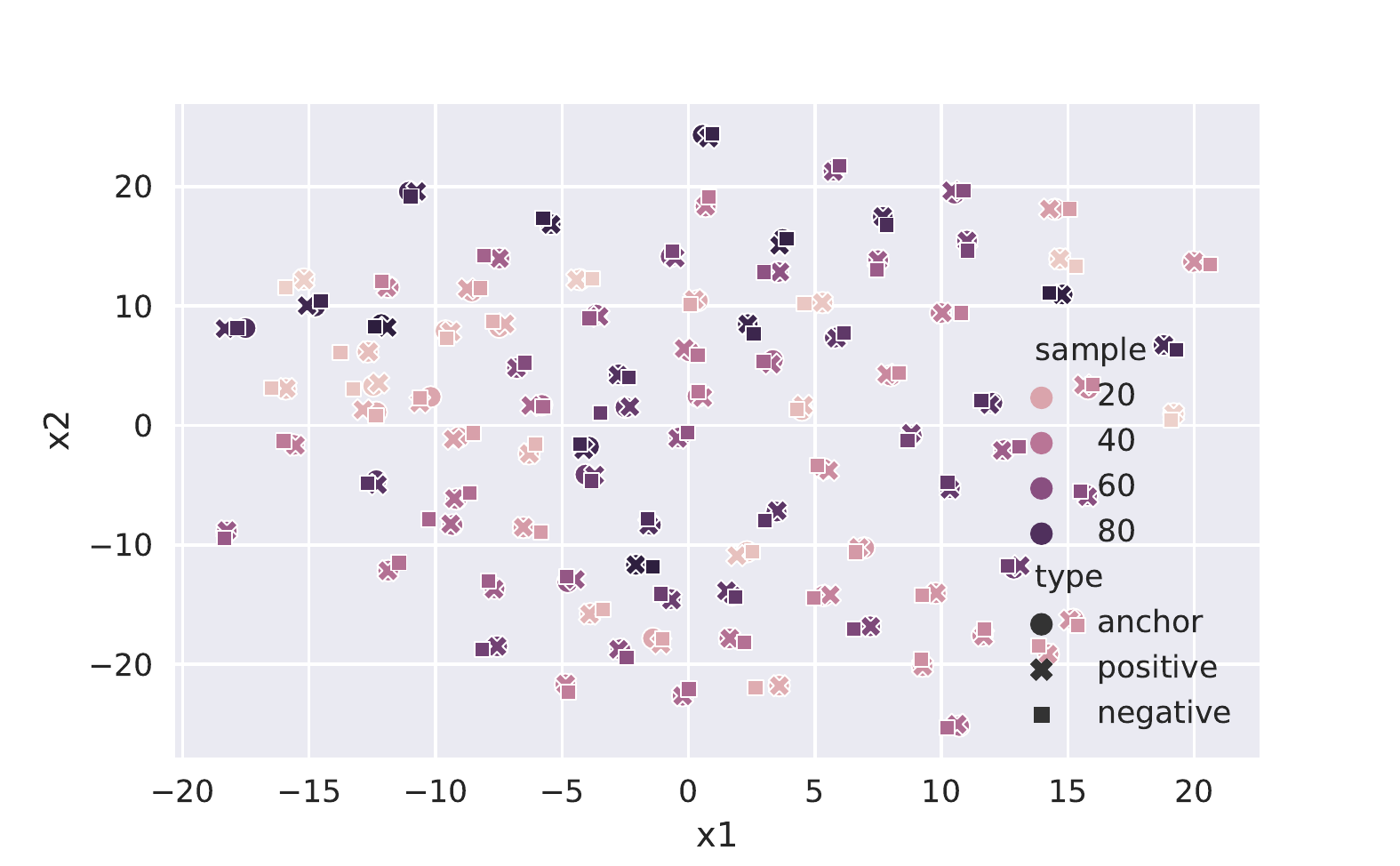}
        \caption{Pretrained}
    \end{subfigure}
% \hspace{0.6em}
\caption{Embedding plots on AquaRAT. Figure  \ref{fig:plots2} in supplementary covers remaining plots.}
\label{fig:plots}
\vspace{-2mm}
\end{figure*}

\begin{figure*}[hbt]
  \begin{subfigure}[b]{\textwidth}
        % \centering
        % \hspace{-45pt}
        \includegraphics[width=0.55\textwidth]{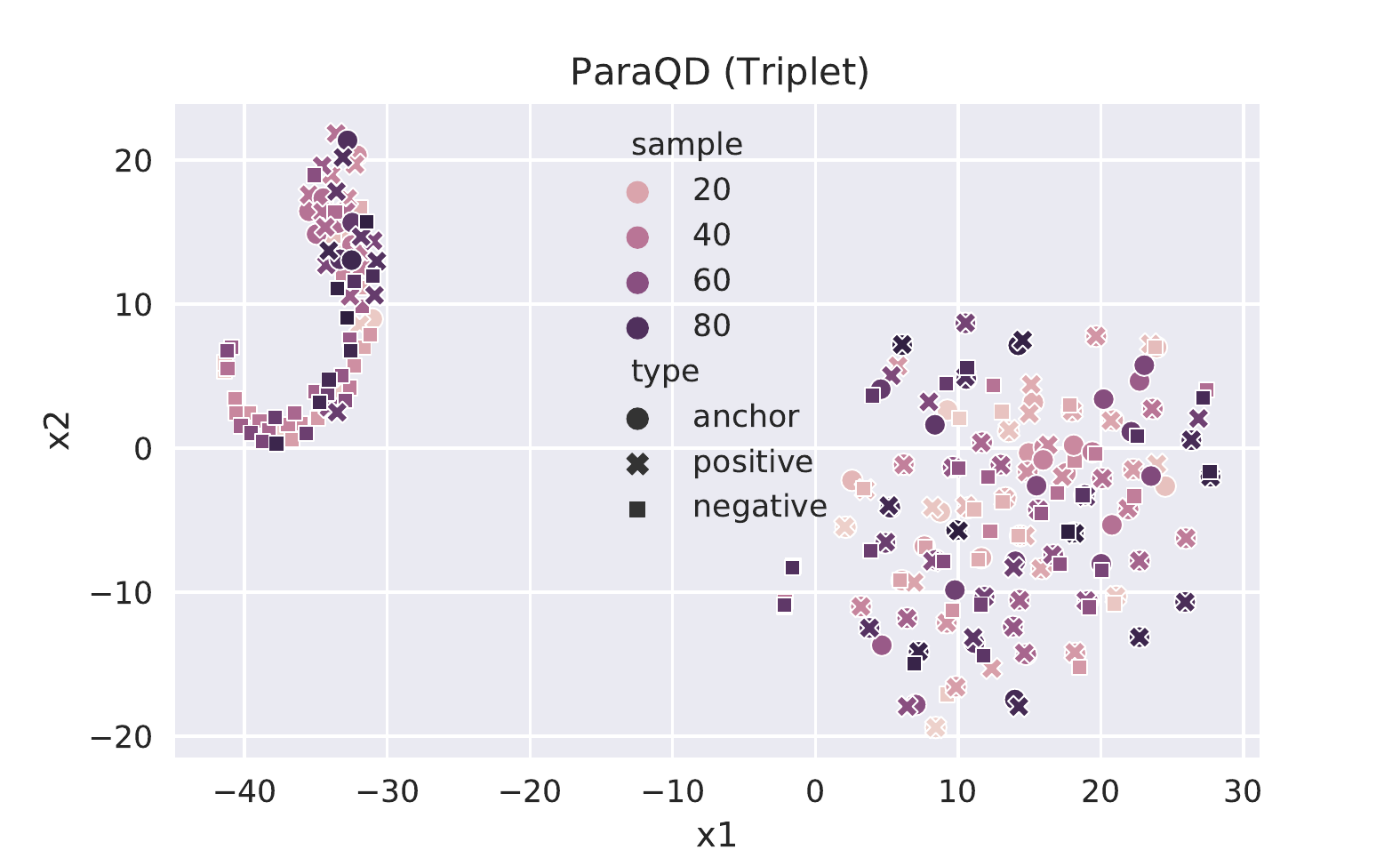}%
        \includegraphics[width=0.55\textwidth]{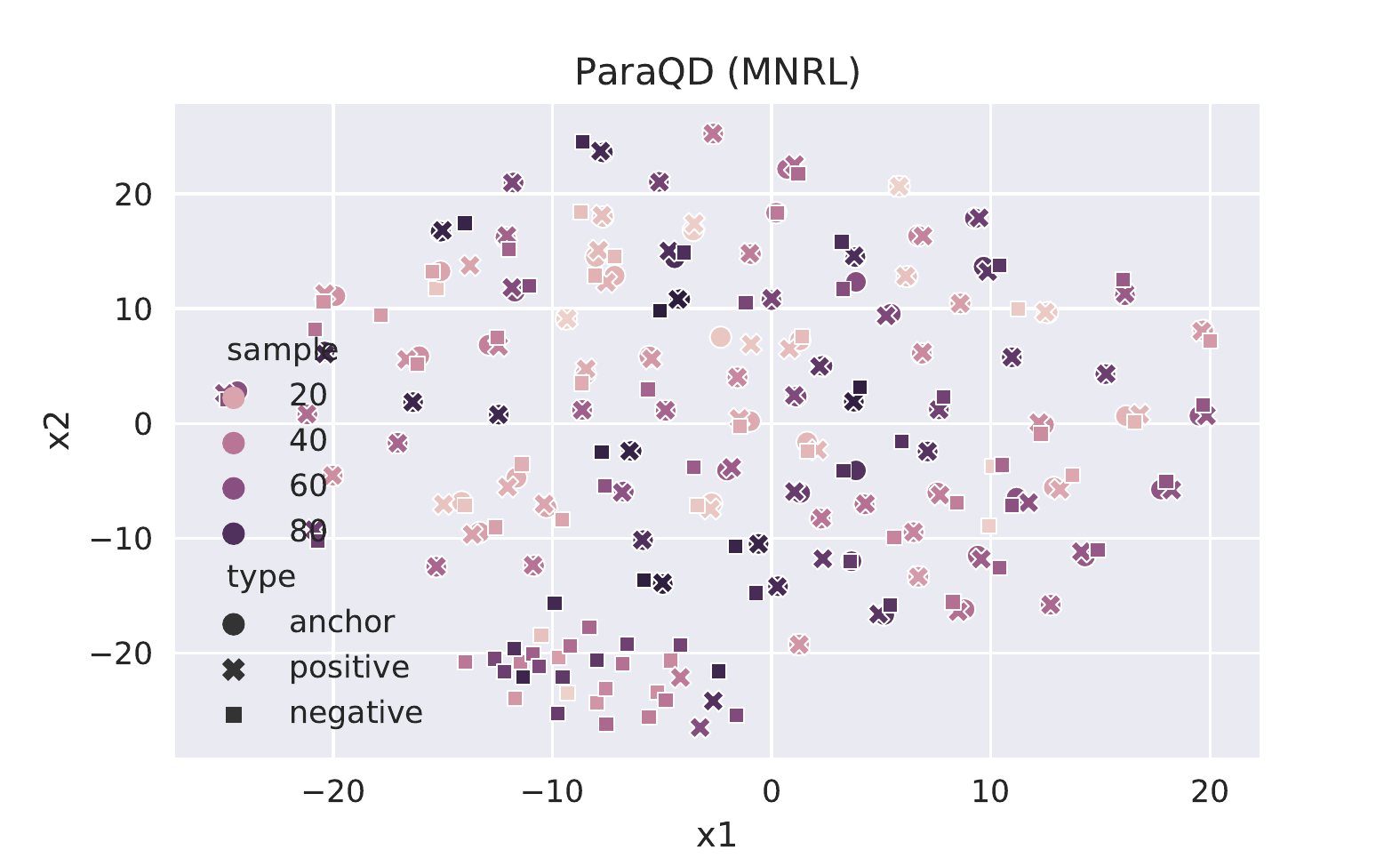}
        % \caption{Loss functions embedding plots comparison}
    \end{subfigure}
\caption{Embedding plots for different loss functions on AquaRAT}
\label{fig:plots-loss}
\end{figure*}

\begin{figure*}[h!]
% \centering
% \hspace{-2pt}
\begin{subfigure}{0.32\linewidth}
%\vspace{-1pt}
\hspace{-2pt}
\includegraphics[width=1.5\linewidth]{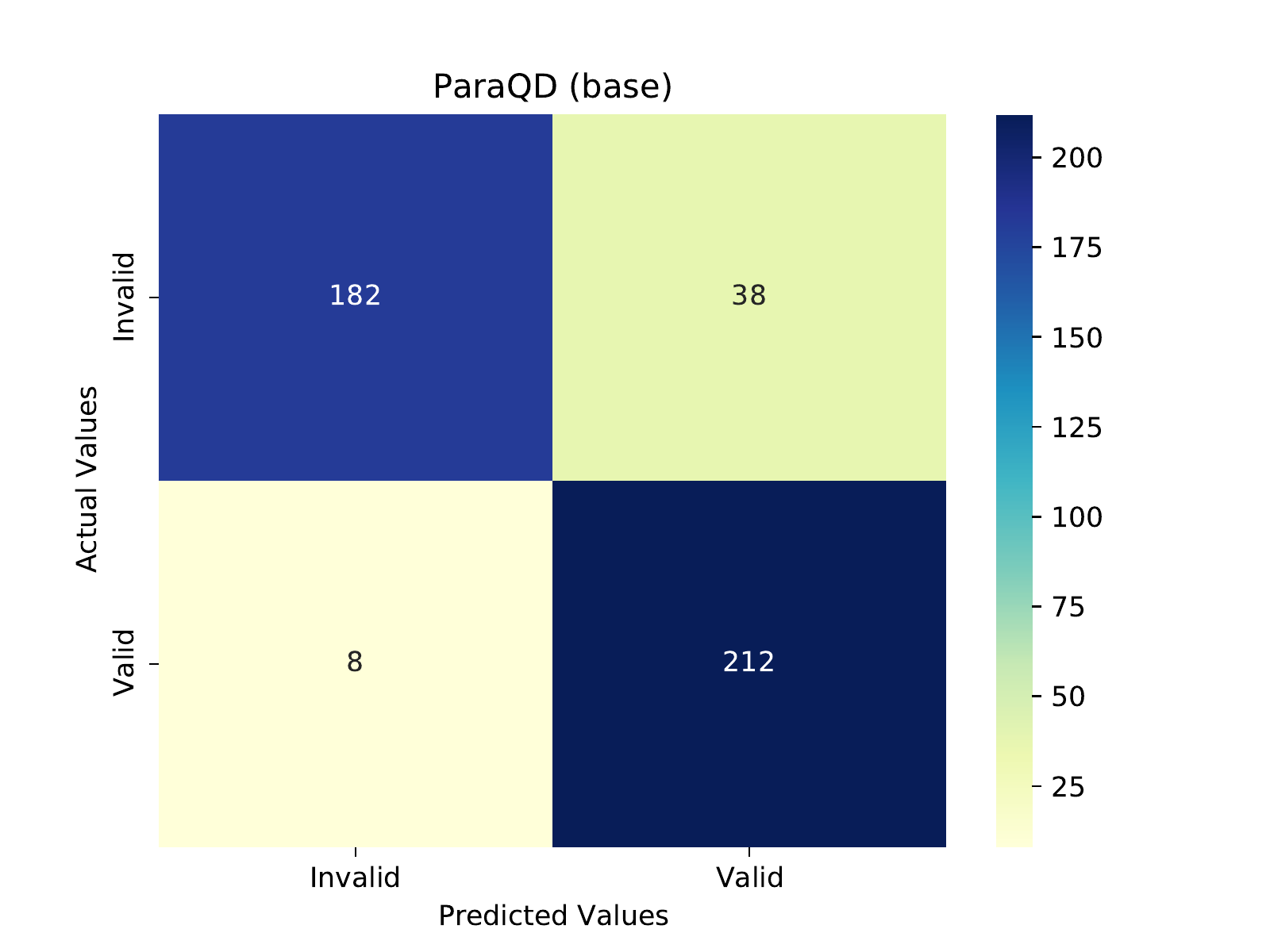}
\caption{ParaQD}
\end{subfigure}
% \hspace{0.3em}
\begin{subfigure}{0.32\linewidth}
%\vspace{-10pt}
\includegraphics[width=1.5\linewidth]{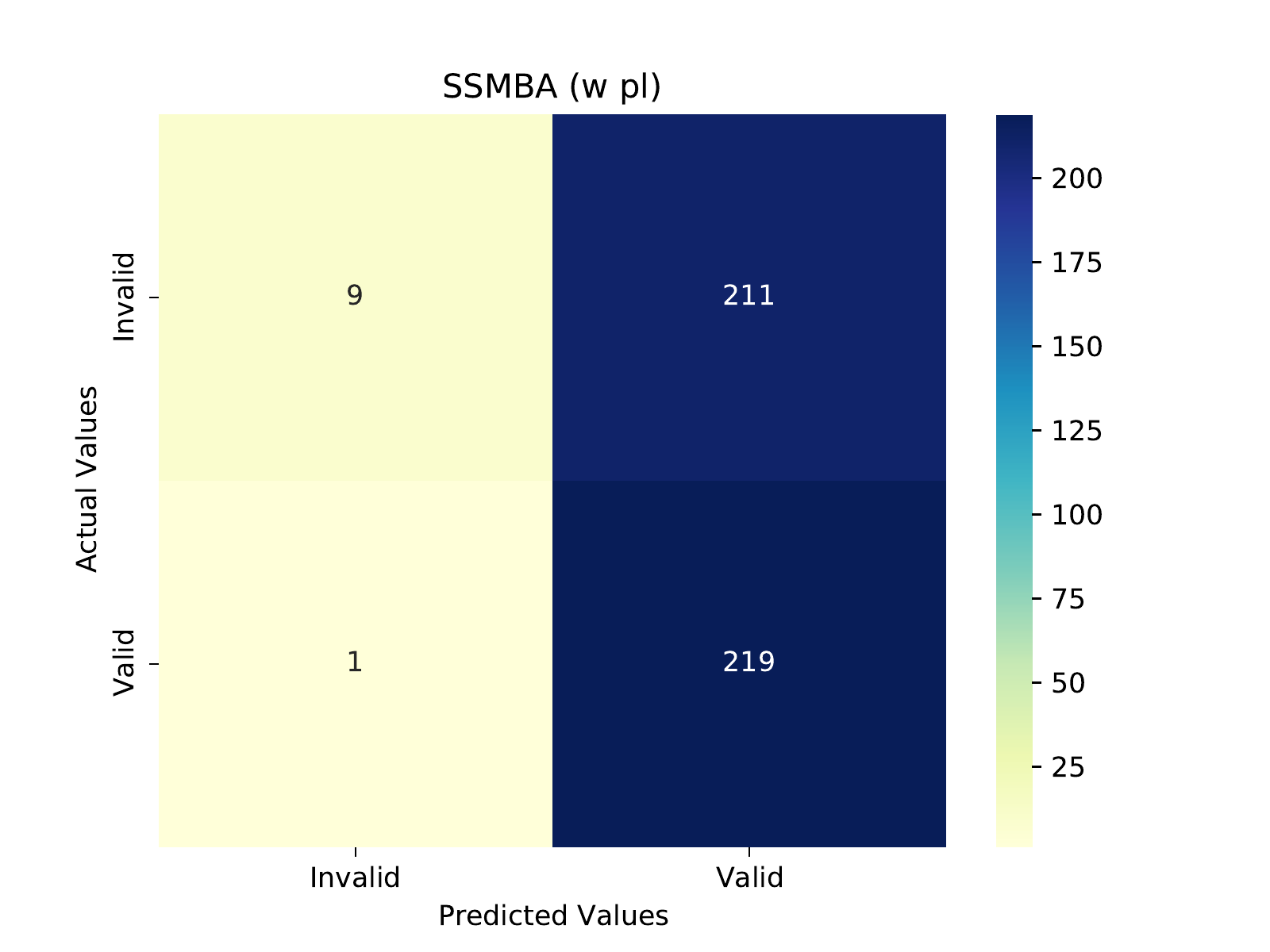}
\caption{SSMBA with PL}
\end{subfigure}%
\hspace{1em}
\begin{subfigure}{0.32\linewidth}
%\vspace{-10pt}
\includegraphics[width=1.5\linewidth]{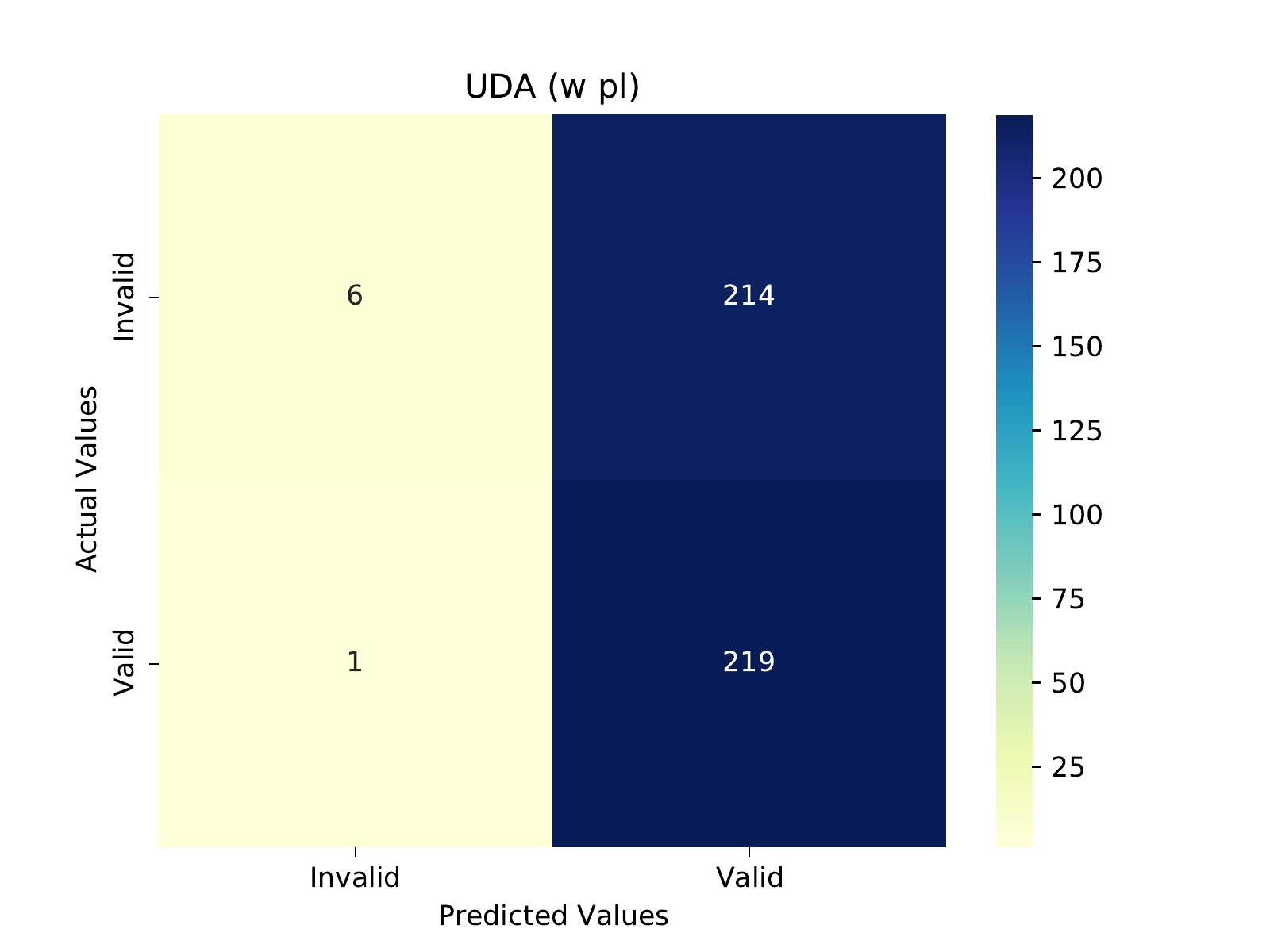}
\caption{UDA w pl}
\end{subfigure}
\label{subfig}
\begin{subfigure}{0.32\linewidth}
%\vspace{-18pt}
\includegraphics[width=1.5\linewidth]{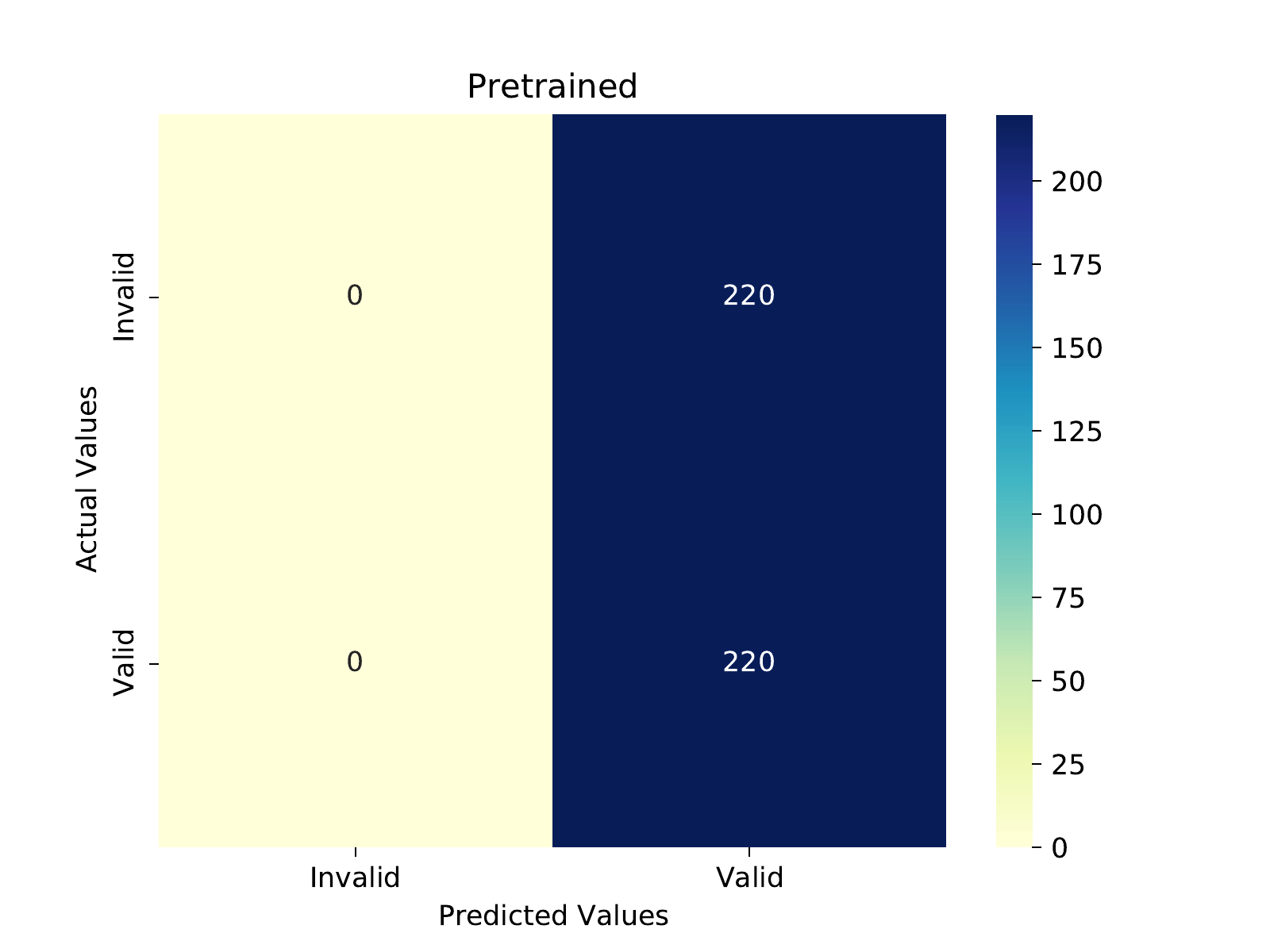}
\caption{Pretrained}
\end{subfigure}%
\begin{subfigure}{0.31\linewidth}
%\vspace{-9pt}
\includegraphics[width=1.5\linewidth]{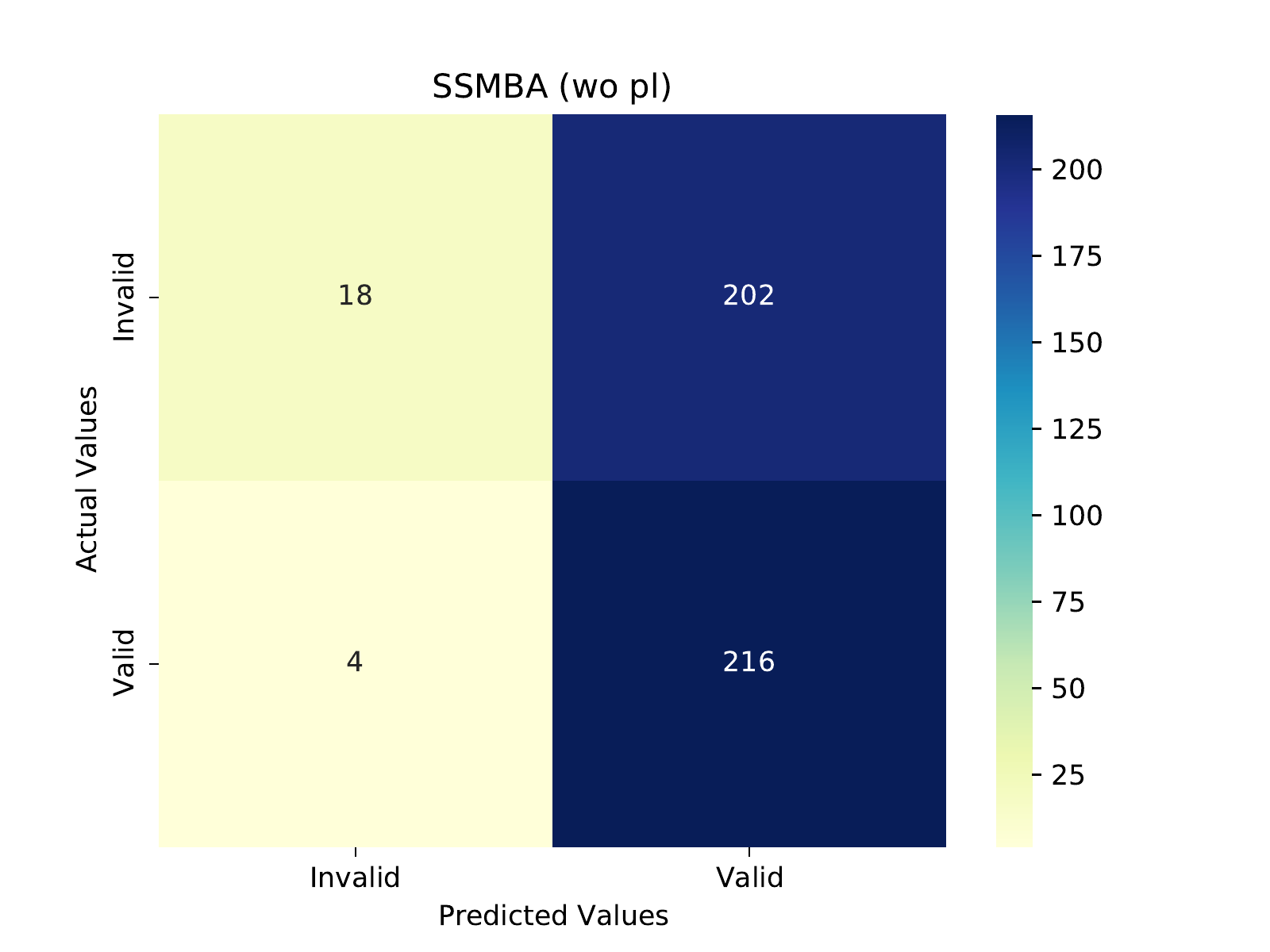}
\caption{SSMBA}
\end{subfigure}
% \hspace{0.6em}
\begin{subfigure}{0.32\linewidth}
%\vspace{-18pt}
\includegraphics[width=1.5\linewidth]{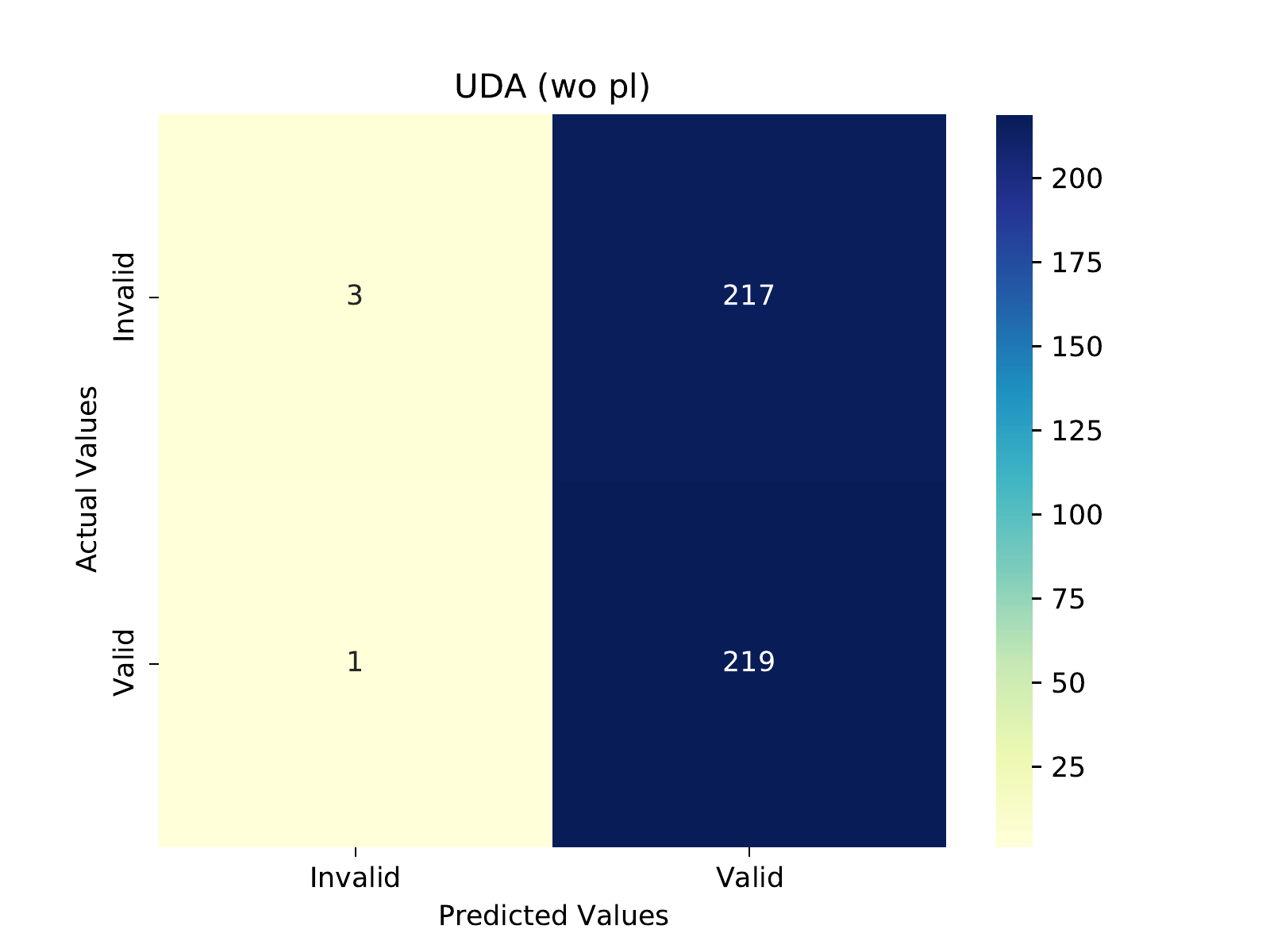}
\caption{UDA}
\end{subfigure}%
% \hspace{1em}
\caption{Confusion matrices for all methods on AquaRAT. Others can be found in supplementary (Figures \ref{fig:cm-SAWP}, \ref{fig:cm-PAWP})}
\label{fig:cm}
\vspace{-2mm}
\end{figure*}

\subsection{Embedding Plots}
To qualitatively evaluate ParaQD, we use t-SNE to project the embeddings into a two-dimensional space (Appendix \ref{appendix:plots}) as seen in Figure \ref{fig:plots}. We observe that the separation between anchors and negatives of triplets is minimal for the baselines, while ParaQD is able to separate them more effectively. Perhaps a more interesting insight from Figure \ref{fig:plots-para} is that our method is able to cluster negatives together, which is not explicitly optimized by triplet loss as it does not account for inter-sample interaction. We note that our negative operators (with the possible exception of $f_7$ and $f_{10}$) are designed to generate unsolvable problems serving as good negatives for training the scoring model (ParaQD). 
% Thus, we hypothesize that the representations learned by our model can be used as a basis for the novel task of \textit{solvable problem detection}. This task entails detecting whether a singular given question is solvable or not. We defer its implementation to future work.
\vspace{-0.2cm}

\subsection{Operator Ablations}
% \vspace{-0.35cm}

To measure the impact of all operators, we trained the model after removing each operator one by one. The summary of the results is in Table \ref{table:top2} (complete in Table \ref{table:ops} (supplementary)). We note that $f_1$ (defined in Section \ref{f1}) seems to be the most consistently important operator amongst the positives, while $f_9$ (defined in Section     \ref{f9}) is the most consistently important operator amongst the negatives. One possible reason for the success of f1 could be that it is the only positive operator that actually changes the words and sentence structure, which is replicated by our test operators and by the human-generated paraphrases. 

Also, for the synthetically generated test sets (for AquaRAT, EM\_Math and SAWP), since $f_9$ is a transformer model, it might generate paraphrases with a closer distribution (especially to $f_e$), but it also performs well on the human crafted paraphrases on PAWP. $f_4$ performs really well on EM\_Math as the dataset involves more mathematical symbols, and thus the distribution of the data is such that technical operators (like $f_4$ and $f_8$) would have a more profound impact on the dataset.  

The results also show that operator importance depends on the data, as certain data distributions might possess patterns that are more suitable to a certain set of operators. We also note that all operators are critical as removing any operator reduces performance for multiple datasets, thus demonstrating the usefulness of the combination of augmentations as a general framework.
\begin{table*}[hbt!]
\small
% \renewcommand{\arraystretch}{1.2}
% \small
\centering
\caption{Analysis of model scores for different examples}

\begin{tabular}{L|L|p{0.8cm}|p{0.9cm}}

% \hline 
& & \\ %\cline{4-7}
 \bf Original & \bf Paraphrase &\bf Label & \bf ParaQD \\ \hline \hline
 A bag of cat food weighs 7 pounds and 4 ounces. How much does the bag weigh in ounces? & A bag of cat food weighs 7 pounds and ounces. How much does the bag in ounces? &0 & -0.922 \\ \hline
 A cart of 20 apples is distributed among 10 students. How much apple does each student get? & 20 hats in a cart are equally distributed among 10 students. How much apple does each student get? & 0 & -0.999 \\\hline
 A cart of 20 apples is distributed among 10 students. How much apple does each student get? & 20 hats in a cart are equally distributed among 10 students. How many hats does each student get? & 1 & 0.999
\\ \hline
John walked 200 kilometres. How long did he walk in terms of metres? & john walked 200 centimetres. How long did he walk in terms of metres? & 0 & -0.999
\\ \hline
John walked 200 kilometres. How long did he walk in terms of metres? & john walked 200 km. How long did he walk in terms of metres? & 1 & 0.999
\\ \hline
% Five dozen gifts are packed in a box and 98 boxes are kept in a tempo. How many tempos can lift 29400 gifts in one round? & Five dozen toys are packed in a box and 98 boxes are kept in a truck. How many trucks can lift 29400 toys in one round? & 1 & -0.999 \\\hline
% Five dozen gifts are packed in a box and 98 boxes are kept in a tempo. How many tempos can lift 29400 gifts in one round? & Five dozen gifts are packed in a box and 98 boxes are kept in a truck. How many trucks can lift 29400 gifts in one round? & 1 & 0.988 \\\hline
\label{tab:results:v}

\end{tabular}
\end{table*}
\subsection{Effects of Loss Functions, Encoder and Seed}
We analyzed the impact of the loss function by performing an ablation with Multiple Negative Ranking Loss (MNRL) (Appendix \ref{appendix:loss}) when training ParaQD. Since MNRL considers inter-sample separation, rather than explicitly distancing the generated hard negative, it is not able to provide a high margin of separation between the positives and negatives ($\mu^s$ = 0.416) as high as the triplet loss ($\mu^s$ = 0.78) but does result in a minor increase in the F1 scores. This can be observed in Figure \ref{fig:plots-loss} and Table \ref{table:loss} (supplementary).
We also analyzed the effects of the encoder and seed across methods on AquaRAT (Table \ref{table:seed}, \ref{table:encoder}; detailed analysis in Appendix \ref{appendix:encoder}) to demonstrate the robustness of our approach. We observe that we outperform the baselines on all the metrics for three encoders we experimented with, namely MiniLM (12 layers), MiniLM (6 layers) and MPNet for different seeds.

\subsection{Error Analysis and Limitations}

\textbf{Does the model check for the preservation of numerical quantities?}:
From example 1 in Table \ref{tab:results:v}, we observe that the number \textbf{4} is missing in the paraphrase rendering the problem unsolvable. Our model outputs a negative score, indicating it is a wrong paraphrase. This general phenomenon is observed in our reported results.
\\
\textbf{Does the model check for entity consistency?}:
We also observe that our model checks for entity consistency. For instance, in example 2, we observe that the paraphraser replaces \textit{apples} with \textit{hats} in the first sentence of the question. However, it fails to replace it in the second part of the question retaining the term \textit{apple} which leads to a low score from ParaQD due to inconsistency. We observe from example 3 that when entity replacement is consistent throughout the question (\textit{apple} replaced by \textit{hats}, the model outputs a high score indicating it is a valid paraphrase. \\
\textbf{Does the model detect changes in units?}: Changing the units in algebraic word problems sometimes may render the question unsolvable or change the existing solution requiring manual intervention. For instance, from example 4 in Table \ref{tab:results:v}, we observe that the unit \textit{kilometres} is changed to \textit{centimetres} in the paraphrase, which would change the equation to solve the question and by consequence the existing solution. Since we prefer solution preserving transformation of the question, ParaQD assigns a low score to this paraphrase. However, when \textit{kilometres} is contracted to \textit{km} in example 5, we observe that our model correctly outputs a high score. \\
\textbf{Does the model make errors under certain scenarios?}:
We also analyzed the errors made by the model. We noted that samples that have valid changes in numbers are not always scored properly by the model. Thus, a limitation of this approach is that it is not robust to changes in numbers that preserve the solution. For instance, if we change the numbers 6 and 4 to 2 and 8 in Figure \ref{fig:example}, the underlying equation and answer would still be preserved. But ParaQD may not output a high score for the same. We must note, however, that generating these types of paraphrases is something that is beyond the ability of general paraphrasing models. As a potential solution (in the future), we propose that numerical changes can be handled through feedback from an automatic word problem solver.
% We defer experimental evaluation of this hypothesis to future work. 
% For instance, in example 6 in Table \ref{tab:resulst:v}, the score assigned by our method to the paraphrase is -0.999, although it is a valid paraphrase. The only changes in this case are the entities \textit{gift} and \textit{tempo} to \textit{toys} and \textit{truck} respectively. However, in the last example, we observe that model is able to correctly output a high score when \textit{tempos} are replaced by \textit{trucks} indicating that \textit{gifts} may have been considered as an important entity by ParaQD resulting in an error in the previous example. 
\section{Conclusion}
In this paper, we formulated the novel task of scoring paraphrases for algebraic questions and proposed a self-supervised method to accomplish this. We demonstrated that the model learns valuable representations that separate positive and negative paraphrases better than existing text augmentation methods and provided a detailed analysis of various components. In the future, we plan to use the scoring model as an objective to steer language models for paraphrasing algebraic word problems and also investigate the usage of representations learned by our method for the novel task of solvable problem detection.

\section{Acknowledgements}
We would sincerely like to thank Extramarks Education India Pvt. Ltd., SERB,
FICCI (PM fellowship) and TiH Anubhuti (IIITD) for supporting this work.

\bibliography{pkdd}
\bibliographystyle{splncs04}

\newpage
\appendix

% \section{Embedding Plots}
% \label{appendix:plots}
% \begin{figure}[h]
%     % \centering
%     \includegraphics[width=2.3\linewidth]{images/AQUA_method_ParaQD.pdf}
%     \caption[Caption]{Embeddings Plot}
%     \label{fig:emb_methods}
% \end{figure}
\begin{table*}[hbt!]
\small
\caption{An ablative analysis of each operator across all datasets. Here, each operator $f_i$ represents the results when we train after removing that operator. The numbers in bold represent the lowest scores for positive operators ($f_1, \dots f_4$) and negative operators ($f_5, \dots f_{10}$) each, thereby demonstrating the impact of that operator.}
\begin{tabular}{clrrrrrrrrp{0.6cm}}
\hline
\multicolumn{1}{l}{\multirow{2}{*}{\textbf{Dataset}}} & \multirow{2}{*}{\textbf{Op}} & \multicolumn{3}{c}{\textbf{Macro}} & \multicolumn{3}{c}{\textbf{Weighted}} & \multicolumn{1}{l}{\multirow{2}{*}{$\mu^+$}} & \multicolumn{1}{l}{\multirow{2}{*}{$\mu^-$}} & \multicolumn{1}{l}{\multirow{2}{*}{$\mu^s$}} \\ \cline{3-8}
\multicolumn{1}{l}{} &  & \multicolumn{1}{l}{P} & \multicolumn{1}{l}{R} & \multicolumn{1}{l}{F1} & \multicolumn{1}{l}{P} & \multicolumn{1}{l}{R} & \multicolumn{1}{l}{F1} & \multicolumn{1}{l}{} & \multicolumn{1}{l}{} & \multicolumn{1}{l}{} \\ \hline
\multirow{10}{*}{AquaRAT} & $f_1$ & 0.667 & 0.681 & 0.674 & 0.749 & 0.611 & 0.673 & 0.742 & 0.024 & 0.718 \\
 & $f_2$ & 0.682 & 0.694 & 0.688 & 0.769 & 0.616 & 0.684 & 0.813 & 0.047 & 0.766 \\
 & $f_3$ & 0.663 & 0.669 & \textbf{0.666} & 0.75 & 0.586 & \textbf{0.658} & 0.783 & 0.115 & \textbf{0.668} \\
 & $f_4$ & 0.679 & 0.686 & 0.682 & 0.768 & 0.602 & 0.675 & 0.827 & 0.084 & 0.744 \\
 & $f_5$ & 0.664 & 0.67 & 0.667 & 0.751 & 0.589 & 0.66 & 0.791 & 0.119 & 0.672 \\
 & $f_6$ & 0.667 & 0.669 & 0.668 & 0.756 & 0.582 & 0.658 & 0.813 & 0.138 & 0.675 \\
 & $f_7$ & 0.667 & 0.678 & 0.673 & 0.753 & 0.602 & 0.669 & 0.771 & 0.069 & 0.702 \\
 & $f_8$ & 0.671 & 0.679 & 0.675 & 0.758 & 0.598 & 0.668 & 0.796 & 0.09 & 0.706 \\
 & $f_9$ & 0.653 & 0.657 & \textbf{0.655} & 0.74 & 0.573 & \textbf{0.646} & 0.77 & 0.145 & \textbf{0.625} \\  
 & $f_{10}$ & 0.678 & 0.687 & 0.683 & 0.766 & 0.607 & 0.677 & 0.813 & 0.078 & 0.735 \\ \hline
\multirow{10}{*}{EM\_Math} & $f_1$ & 0.635 & 0.644 & 0.64 & 0.67 & 0.627 & 0.648 & 0.425 & -0.144 & 0.569 \\
 & $f_2$ & 0.648 & 0.651 & 0.65 & 0.689 & 0.613 & 0.649 & 0.598 & -0.005 & 0.603 \\
 & $f_3$ & 0.666 & 0.669 & 0.667 & 0.709 & 0.628 & 0.666 & 0.661 & -0.017 & 0.678 \\
 & $f_4$ & 0.638 & 0.635 & \textbf{0.636} & 0.681 & 0.588 & \textbf{0.631} & 0.633 & 0.096 & \textbf{0.538} \\
 & $f_5$ & 0.653 & 0.653 & 0.653 & 0.697 & 0.608 & 0.65 & 0.651 & 0.042 & 0.609 \\
 & $f_6$ & 0.657 & 0.652 & 0.655 & 0.703 & 0.602 & 0.648 & 0.696 & 0.088 & 0.608 \\
 & $f_7$ & 0.66 & 0.658 & 0.659 & 0.705 & 0.612 & 0.655 & 0.681 & 0.048 & 0.633 \\
 & $f_8$ & 0.646 & 0.648 & \textbf{0.647} & 0.688 & 0.608 & 0.646 & 0.606 & 0.018 & \textbf{0.588} \\
 & $f_9$ & 0.656 & 0.649 & 0.652 & 0.702 & 0.597 & \textbf{0.645} & 0.704 & 0.108 & 0.596 \\ 
 & $f_{10}$ & 0.672 & 0.674 & 0.673 & 0.716 & 0.632 & 0.671 & 0.677 & -0.012 & 0.689 \\ \hline
\multirow{10}{*}{SAWP} & $f_1$ & 0.618 & 0.627 & 0.623 & 0.688 & 0.572 & 0.625 & 0.572 & 0.06 & 0.512 \\
 & $f_2$ & 0.617 & 0.621 & \textbf{0.619} & 0.69 & 0.552 & \textbf{0.614} & 0.631 & 0.152 & \textbf{0.479} \\
 & $f_3$ & 0.624 & 0.63 & 0.627 & 0.697 & 0.565 & 0.624 & 0.632 & 0.114 & 0.518 \\
 & $f_4$ & 0.646 & 0.648 & 0.647 & 0.725 & 0.57 & 0.638 & 0.739 & 0.16 & 0.579 \\
 & $f_5$ & 0.648 & 0.644 & 0.646 & 0.729 & 0.56 & 0.633 & 0.777 & 0.205 & 0.572 \\
 & $f_6$ & 0.629 & 0.632 & 0.631 & 0.705 & 0.56 & 0.624 & 0.679 & 0.148 & 0.53 \\
 & $f_7$ & 0.63 & 0.637 & 0.634 & 0.704 & 0.572 & 0.632 & 0.649 & 0.096 & 0.553 \\
 & $f_8$ & 0.643 & 0.64 & 0.642 & 0.723 & 0.558 & 0.63 & 0.754 & 0.195 & 0.559 \\
 & $f_9$ & 0.626 & 0.622 & \textbf{0.624} & 0.705 & 0.538 & \textbf{0.61} & 0.723 & 0.245 & \textbf{0.478} \\ 
 & $f_{10}$ & 0.645 & 0.642 & 0.643 & 0.725 & 0.56 & 0.632 & 0.763 & 0.2 & 0.562 \\ \hline
\multirow{10}{*}{PAWP} & $f_1$ & 0.644 & 0.623 & \textbf{0.634} & 0.645 & 0.622 & \textbf{0.633} & 0.64 & 0.148 & \textbf{0.492} \\
 & $f_2$ & 0.68 & 0.636 & 0.658 & 0.681 & 0.635 & 0.657 & 0.769 & 0.226 & 0.543 \\
 & $f_3$ & 0.712 & 0.659 & 0.684 & 0.712 & 0.658 & 0.684 & 0.819 & 0.192 & 0.627 \\
 & $f_4$ & 0.714 & 0.654 & 0.683 & 0.714 & 0.652 & 0.682 & 0.847 & 0.224 & 0.623 \\
 & $f_5$ & 0.694 & 0.644 & 0.668 & 0.695 & 0.642 & 0.668 & 0.8 & 0.229 & 0.57 \\
 & $f_6$ & 0.705 & 0.646 & 0.674 & 0.706 & 0.645 & 0.674 & 0.829 & 0.244 & 0.585 \\
 & $f_7$ & 0.698 & 0.651 & 0.674 & 0.698 & 0.65 & 0.673 & 0.795 & 0.185 & 0.61 \\
 & $f_8$ & 0.702 & 0.629 & 0.663 & 0.702 & 0.628 & 0.663 & 0.859 & 0.343 & 0.516 \\
 & $f_9$ & 0.663 & 0.619 & 0.64 & 0.663 & 0.618 & 0.64 & 0.759 & 0.284 & 0.475 \\ 
 & $f_{10}$ & 0.655 & 0.587 &\textbf{0.619} & 0.656 & 0.585 & \textbf{0.618} & 0.845 & 0.493 & \textbf{0.352} \\ \hline
 \vspace{-2mm}
\end{tabular}

\label{table:ops}
\end{table*}
\begin{figure*}[hbt!]
% \centering
% \hspace{-2pt}
   \begin{subfigure}[b]{\textwidth}
        \centering
        \includegraphics[width=0.5\linewidth]{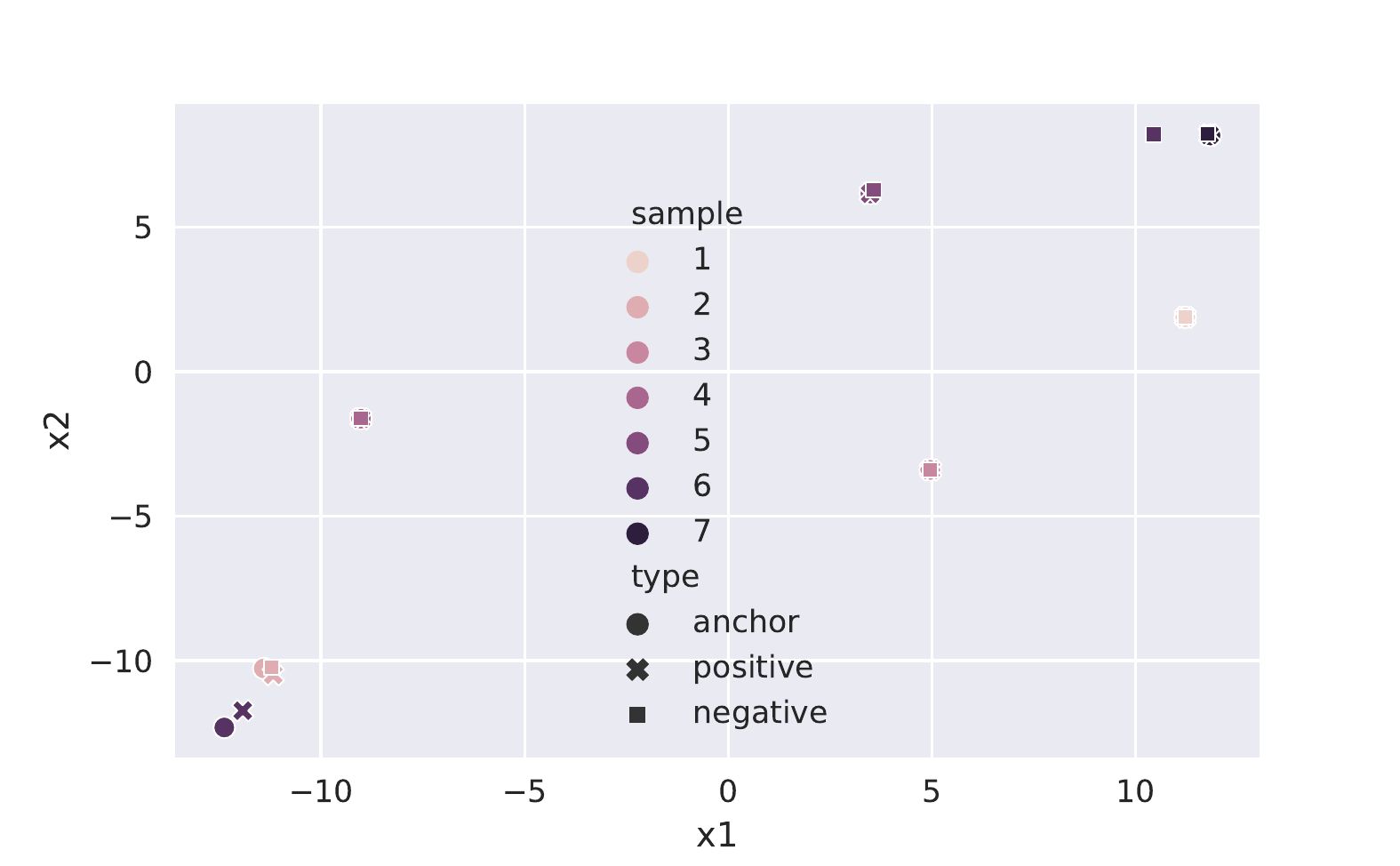}%
        \includegraphics[width=0.5\linewidth]{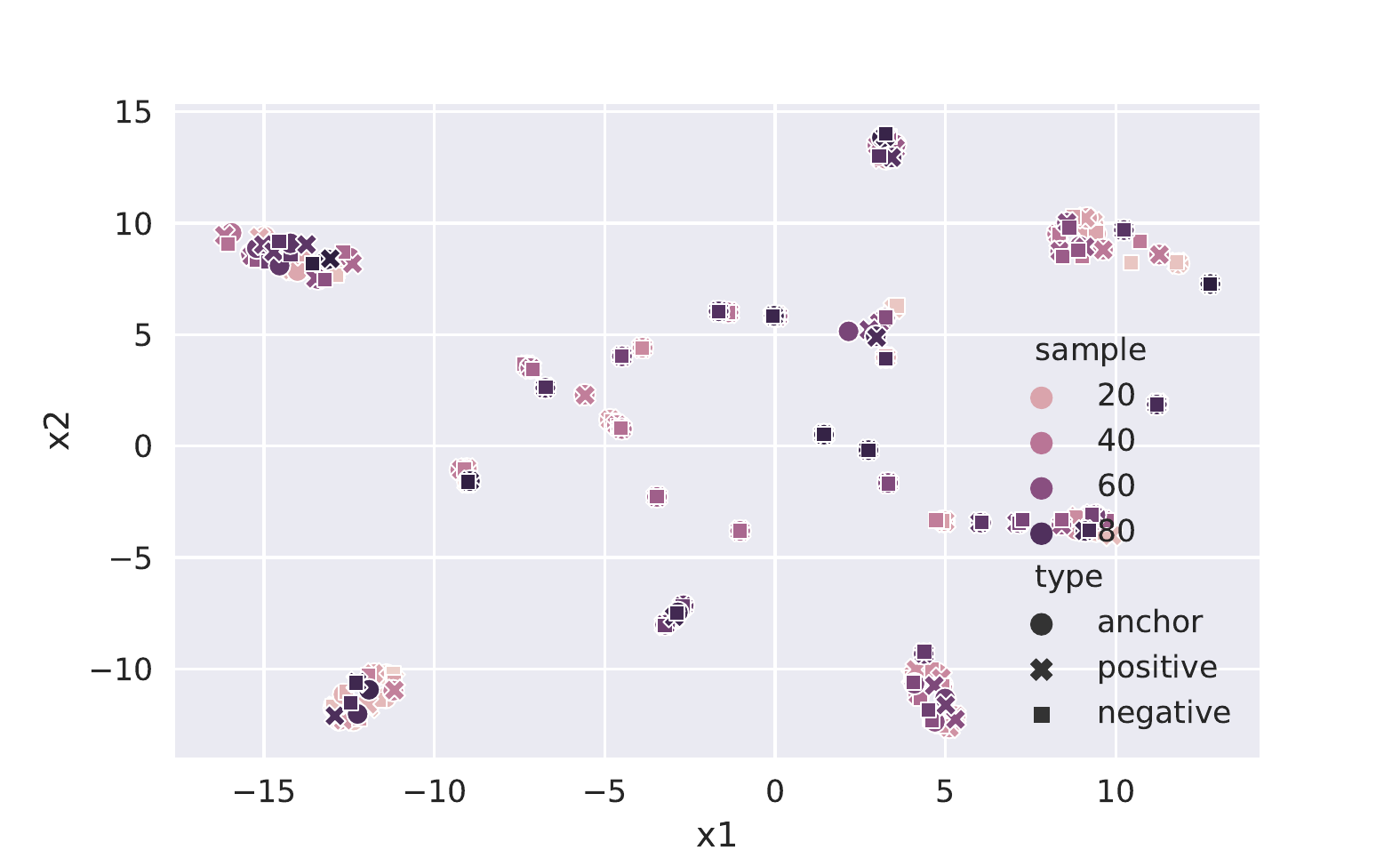}
        \caption{UDA}
    \end{subfigure}
   \begin{subfigure}[b]{\textwidth}
        \centering
        \includegraphics[width=0.49\linewidth]{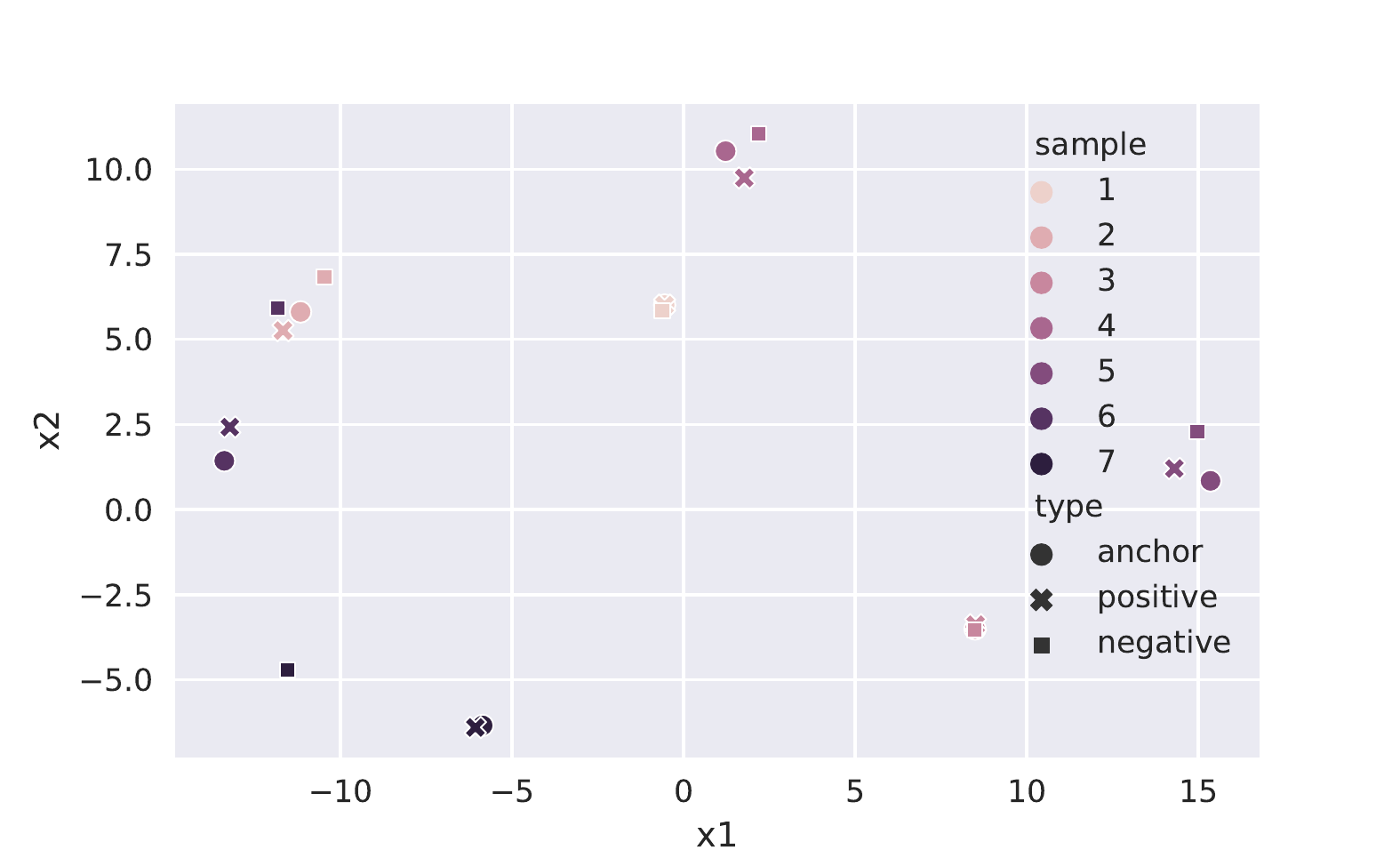}
        \includegraphics[width=0.49\linewidth]{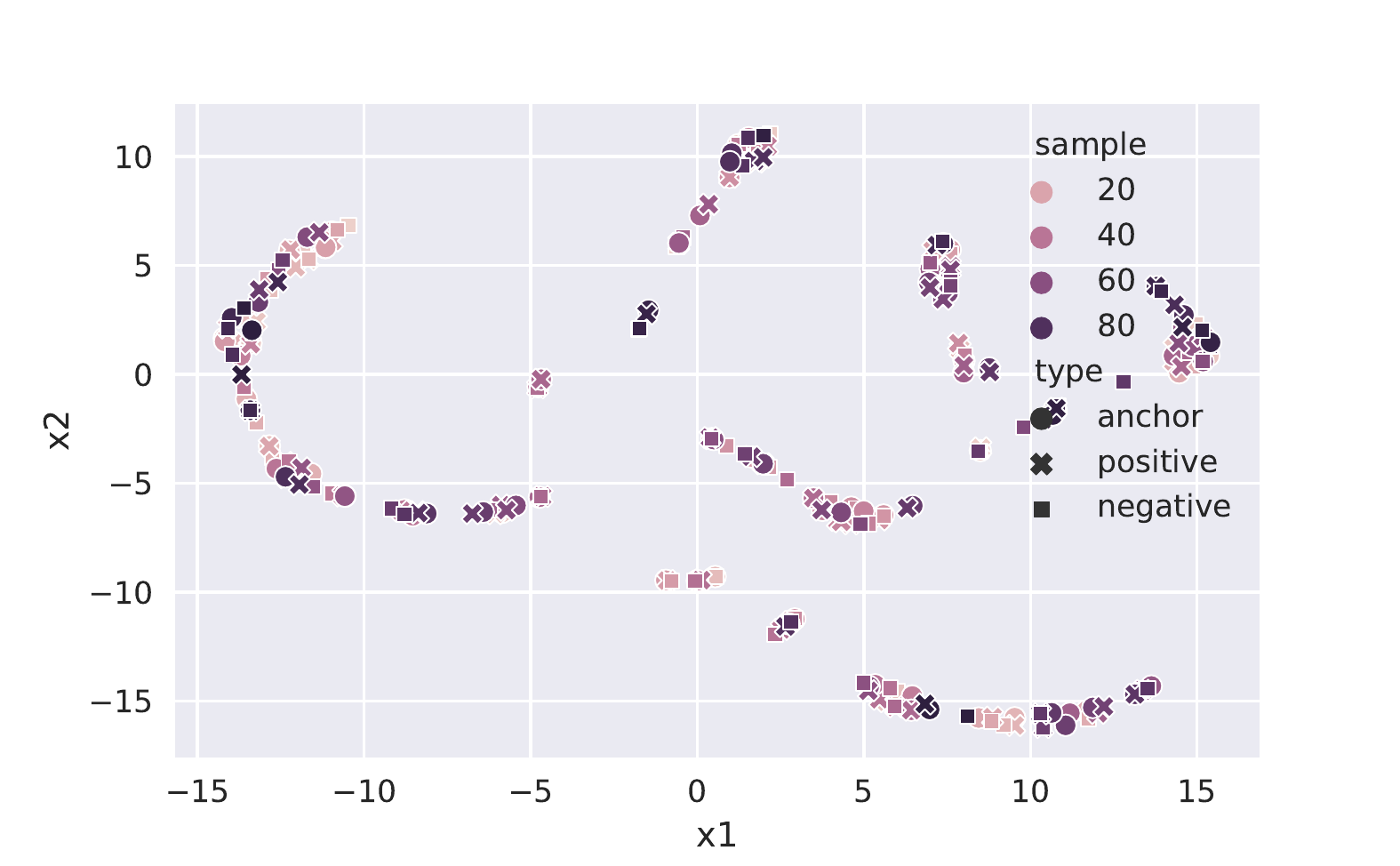}
        \caption{SSMBA}
    \end{subfigure}
% \hspace{0.6em}
\caption{Further Embedding plots on AquaRAT}
\label{fig:plots2}
\vspace{-2mm}
\end{figure*}

\begin{figure*}[h!]
\begin{subfigure}{0.32\linewidth}
\hspace{-1.2cm}
\includegraphics[width=1.25\linewidth]{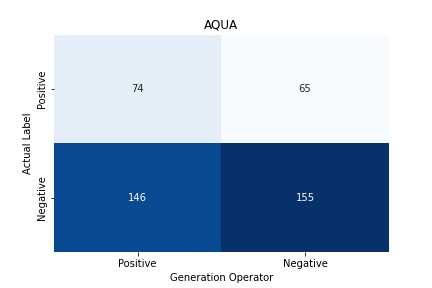}
\caption{AquaRAT}
\end{subfigure}
\begin{subfigure}{0.32\linewidth}
\includegraphics[width=1.25\linewidth]{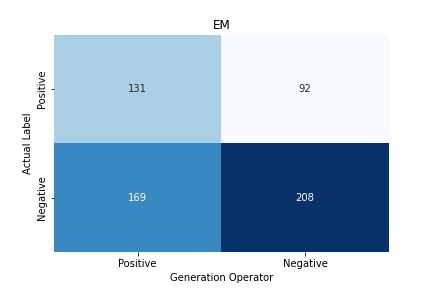}
\caption{EM}
\end{subfigure}
% \hspace{1em}
\begin{subfigure}{0.32\linewidth}
\includegraphics[width=1.25\linewidth]{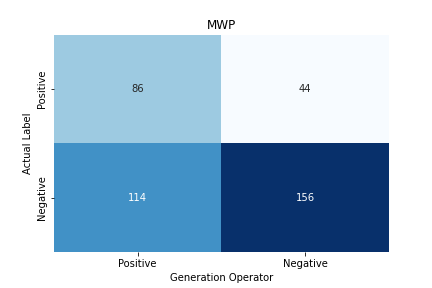}
\caption{SAWP}
\end{subfigure}
\caption{Statistics for Test Operators for AquaRAT, EM and SAWP.}
\label{fig:test}
% \vspace{-2mm}
\end{figure*}

\begin{figure*}[h!]
% \centering
% \hspace{-2pt}
\begin{subfigure}{0.32\linewidth}
%\vspace{-1pt}
\hspace{-2pt}
\includegraphics[width=1.5\linewidth]{images/confusion_matrices/AQUA/ParaQD__base_.pdf}
\caption{ParaQD}
\end{subfigure}
% \hspace{0.3em}
\begin{subfigure}{0.32\linewidth}
%\vspace{-10pt}
\includegraphics[width=1.5\linewidth]{images/confusion_matrices/AQUA/SSMBA__w_pl_.pdf}
\caption{SSMBA with PL}
\end{subfigure}%
\hspace{1em}
\begin{subfigure}{0.32\linewidth}
%\vspace{-10pt}
\includegraphics[width=1.5\linewidth]{images/confusion_matrices/AQUA/UDA__w_pl_.pdf}
\caption{UDA w pl}
\end{subfigure}
\label{subfig}
\begin{subfigure}{0.32\linewidth}
%\vspace{-18pt}
\includegraphics[width=1.5\linewidth]{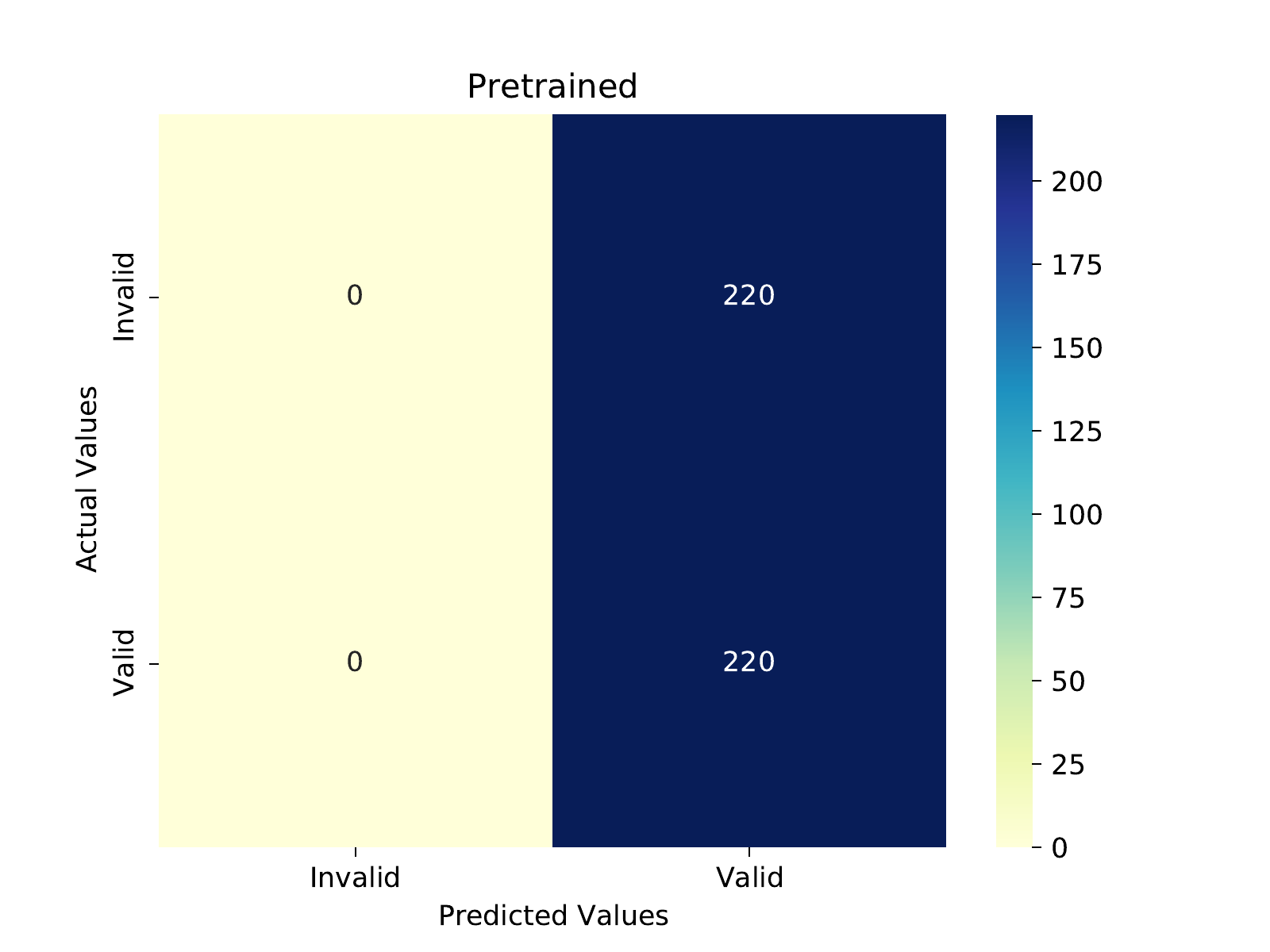}
\caption{Pretrained}
\end{subfigure}%
\begin{subfigure}{0.31\linewidth}
%\vspace{-9pt}
\includegraphics[width=1.5\linewidth]{images/confusion_matrices/AQUA/SSMBA__wo_pl_.pdf}
\caption{SSMBA}
\end{subfigure}
% \hspace{0.6em}
\begin{subfigure}{0.32\linewidth}
%\vspace{-18pt}
\includegraphics[width=1.5\linewidth]{images/confusion_matrices/AQUA/UDA__wo_pl_.pdf}
\caption{UDA}
\end{subfigure}%
% \hspace{1em}
\caption{Confusion matrices for all methods on AquaRAT}
\label{fig:cm}
\vspace{-2mm}
\end{figure*}

\begin{figure*}[hbt!]
% \centering
% \hspace{-2pt}
\begin{subfigure}{0.32\linewidth}
%\vspace{-1pt}
\hspace{-2pt}
\includegraphics[width=1.5\linewidth]{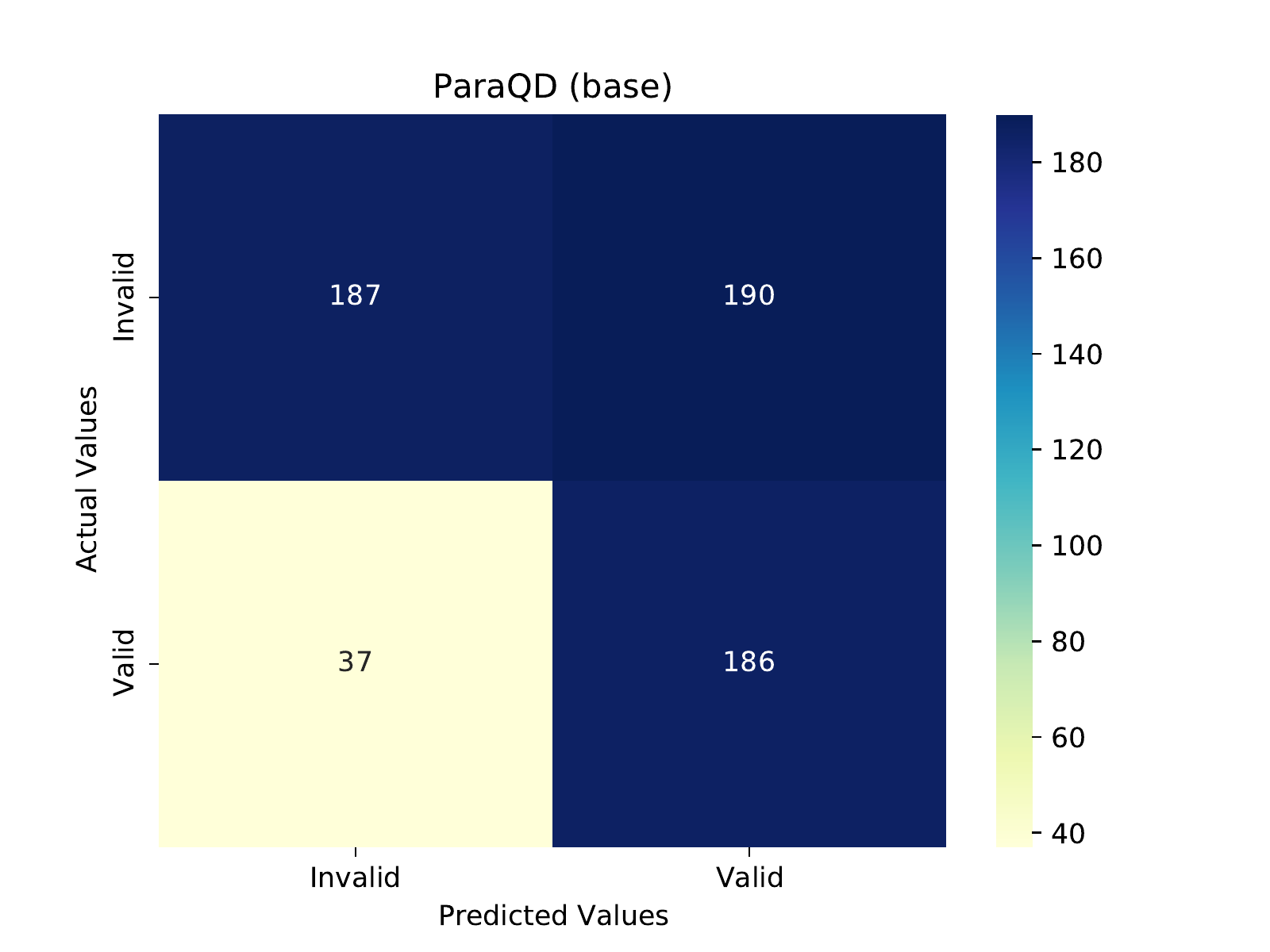}
\caption{ParaQD}
\end{subfigure}
% \hspace{0.3em}
\begin{subfigure}{0.32\linewidth}
%\vspace{-10pt}
\includegraphics[width=1.5\linewidth]{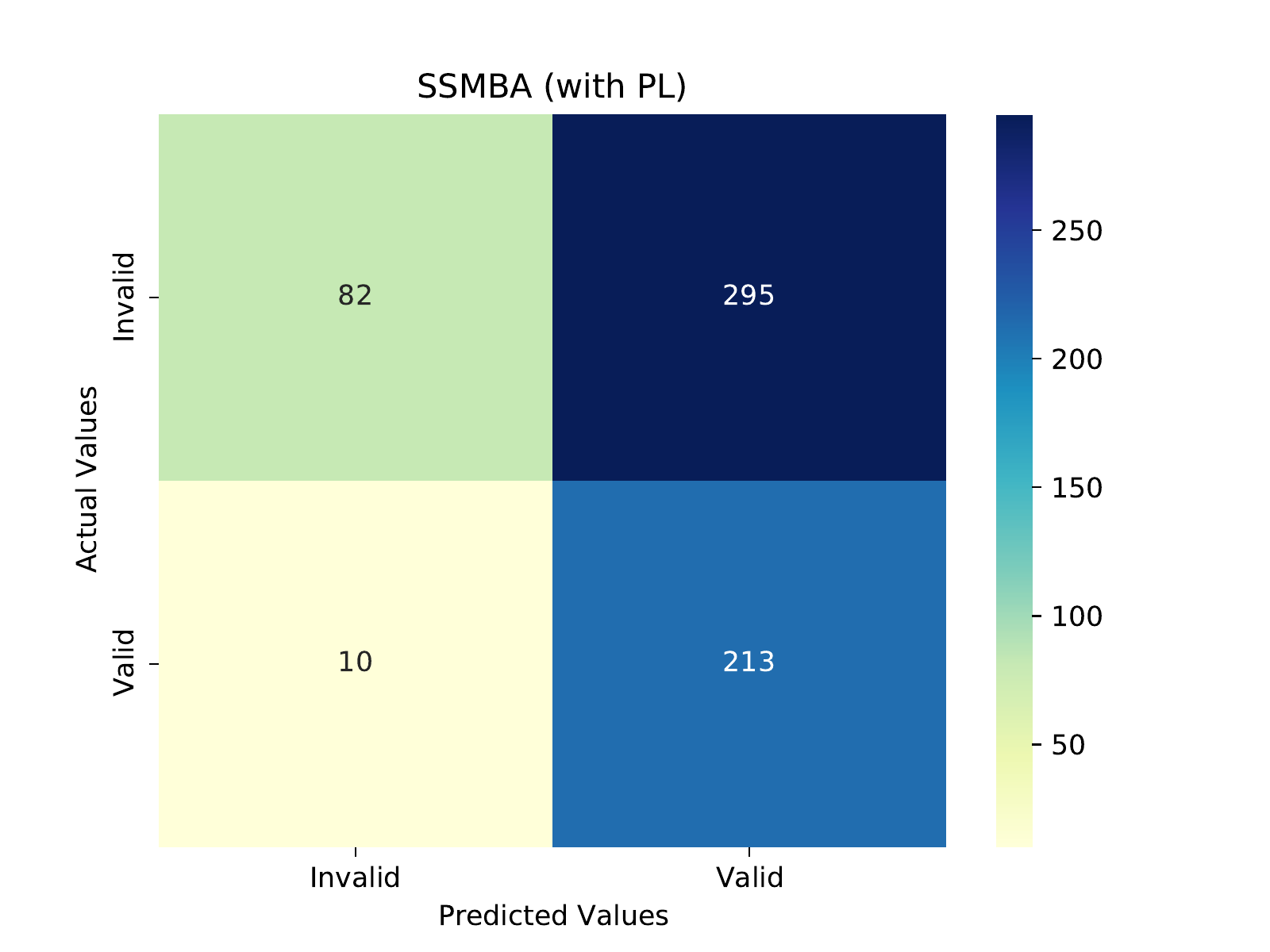}
\caption{SSMBA with pl}
\end{subfigure}%
\hspace{1em}
\begin{subfigure}{0.32\linewidth}
%\vspace{-10pt}
\includegraphics[width=1.5\linewidth]{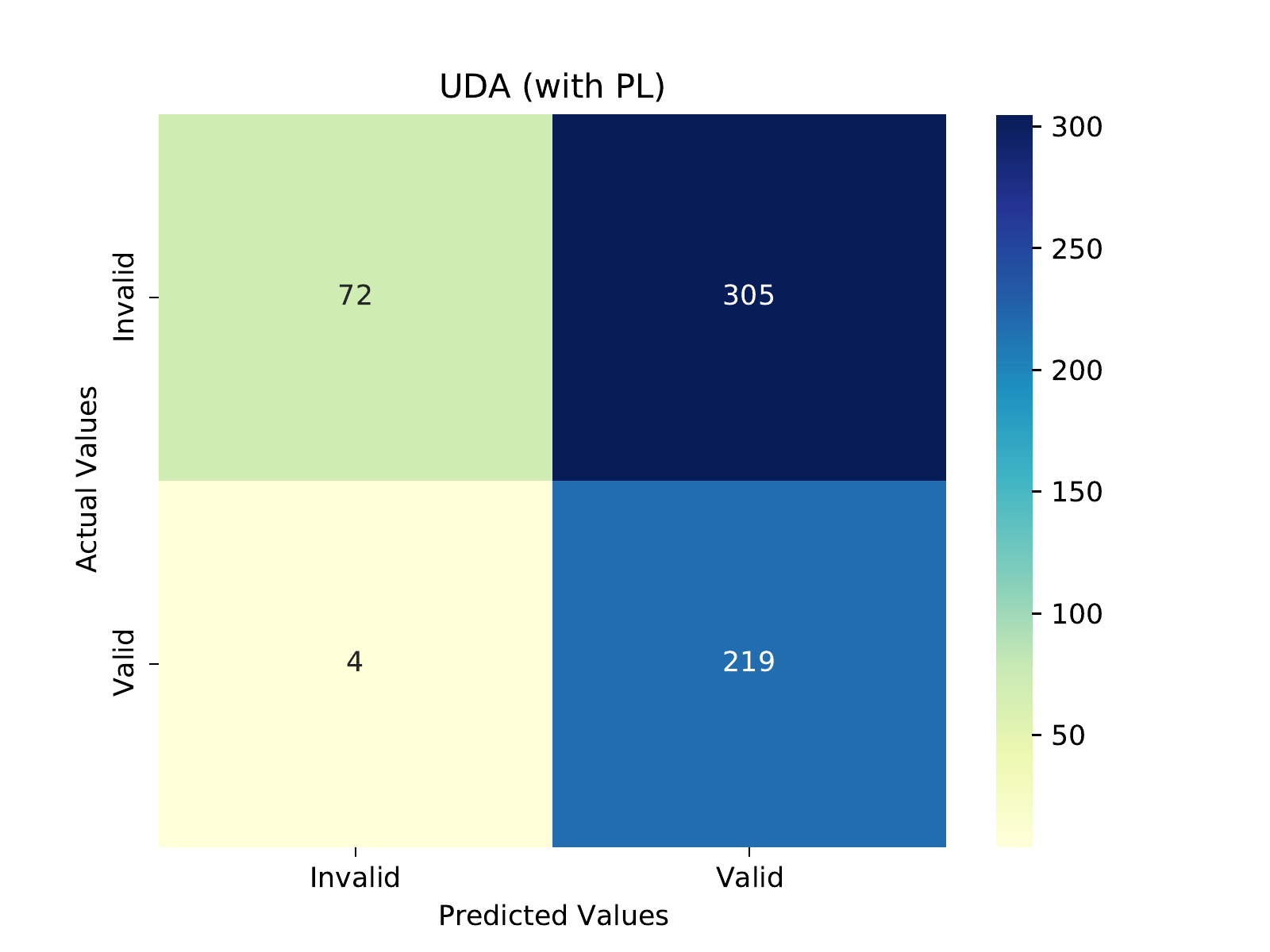}
\caption{UDA with pl}
\end{subfigure}
\label{subfig}
\begin{subfigure}{0.32\linewidth}
%\vspace{-18pt}
\includegraphics[width=1.5\linewidth]{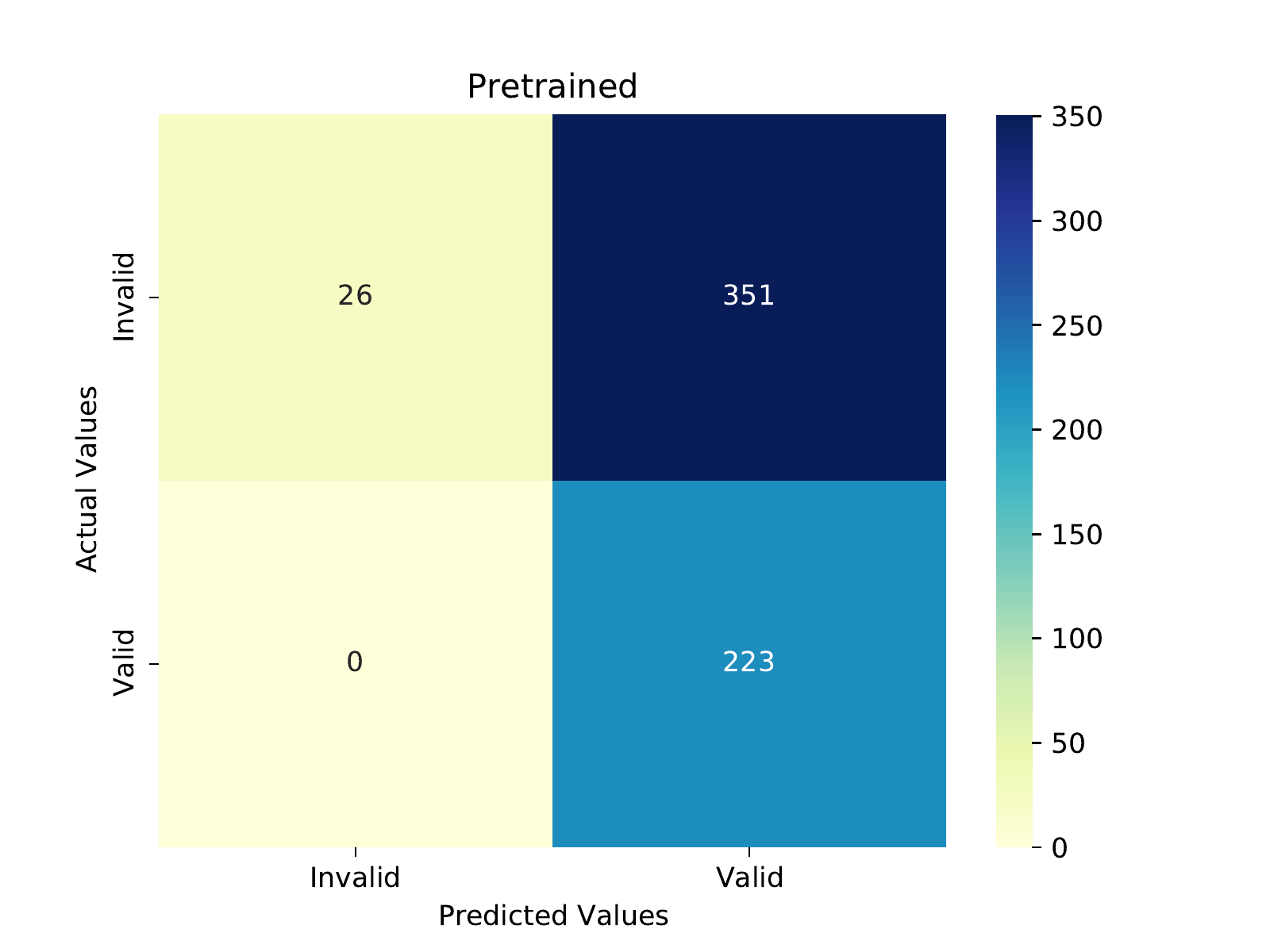}
\caption{Pretrained}
\end{subfigure}%
\begin{subfigure}{0.31\linewidth}
%\vspace{-9pt}
\includegraphics[width=1.5\linewidth]{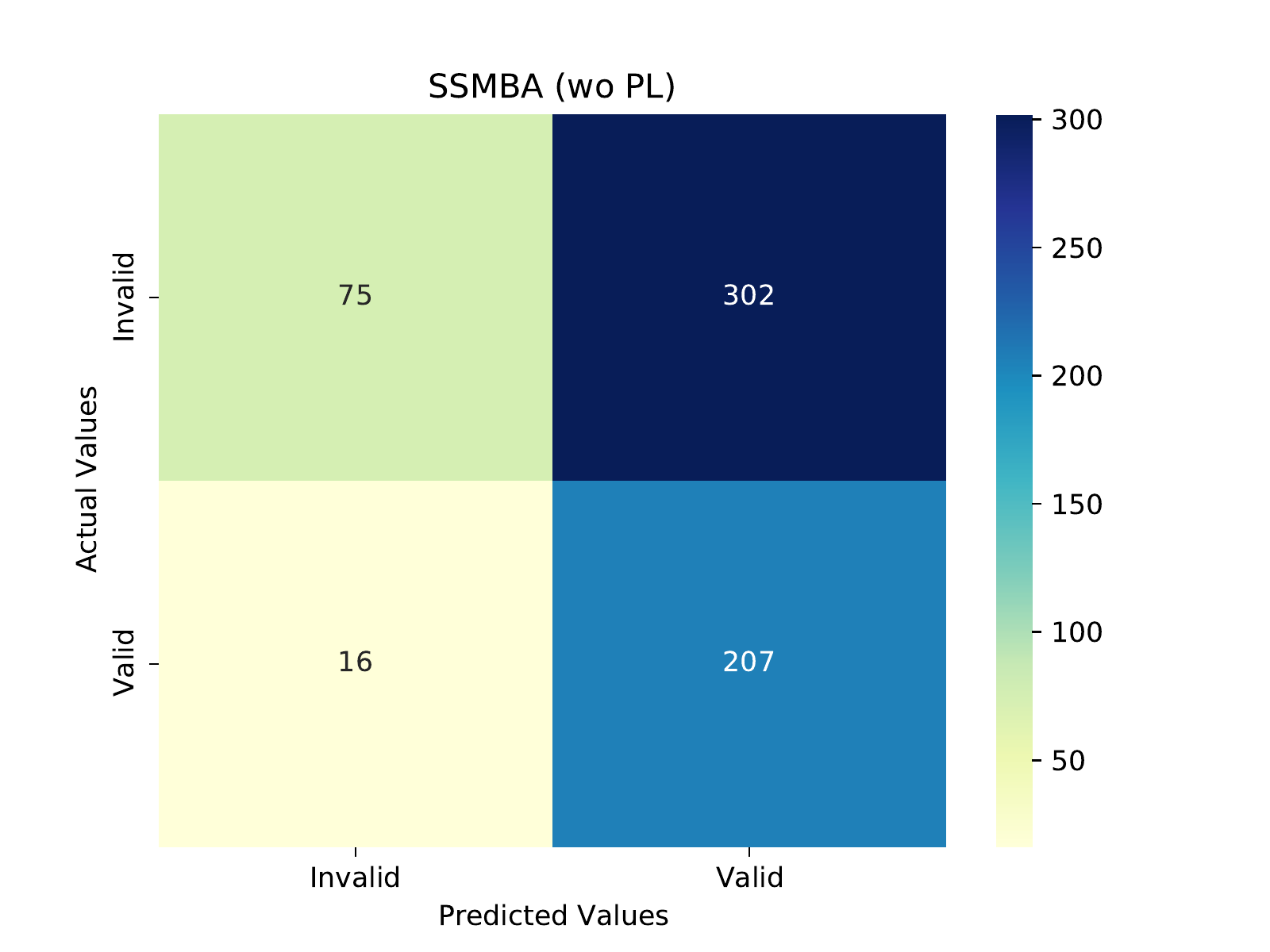}
\caption{SSMBA}
\end{subfigure}
% \hspace{0.6em}
\begin{subfigure}{0.32\linewidth}
%\vspace{-18pt}
\includegraphics[width=1.5\linewidth]{images/confusion_matrices/EM/UDA__wo_pl_.pdf}
\caption{UDA}
\end{subfigure}%
\hspace{1em}
\caption{Confusion matrices for all methods on EM\_Math.}
\label{fig:cm-EM}
\vspace{-2mm}
\end{figure*}

\begin{figure*}[hbt!]
% \centering
% \hspace{-2pt}
\begin{subfigure}{0.32\linewidth}
%\vspace{-1pt}
\hspace{-2pt}
\includegraphics[width=1.5\linewidth]{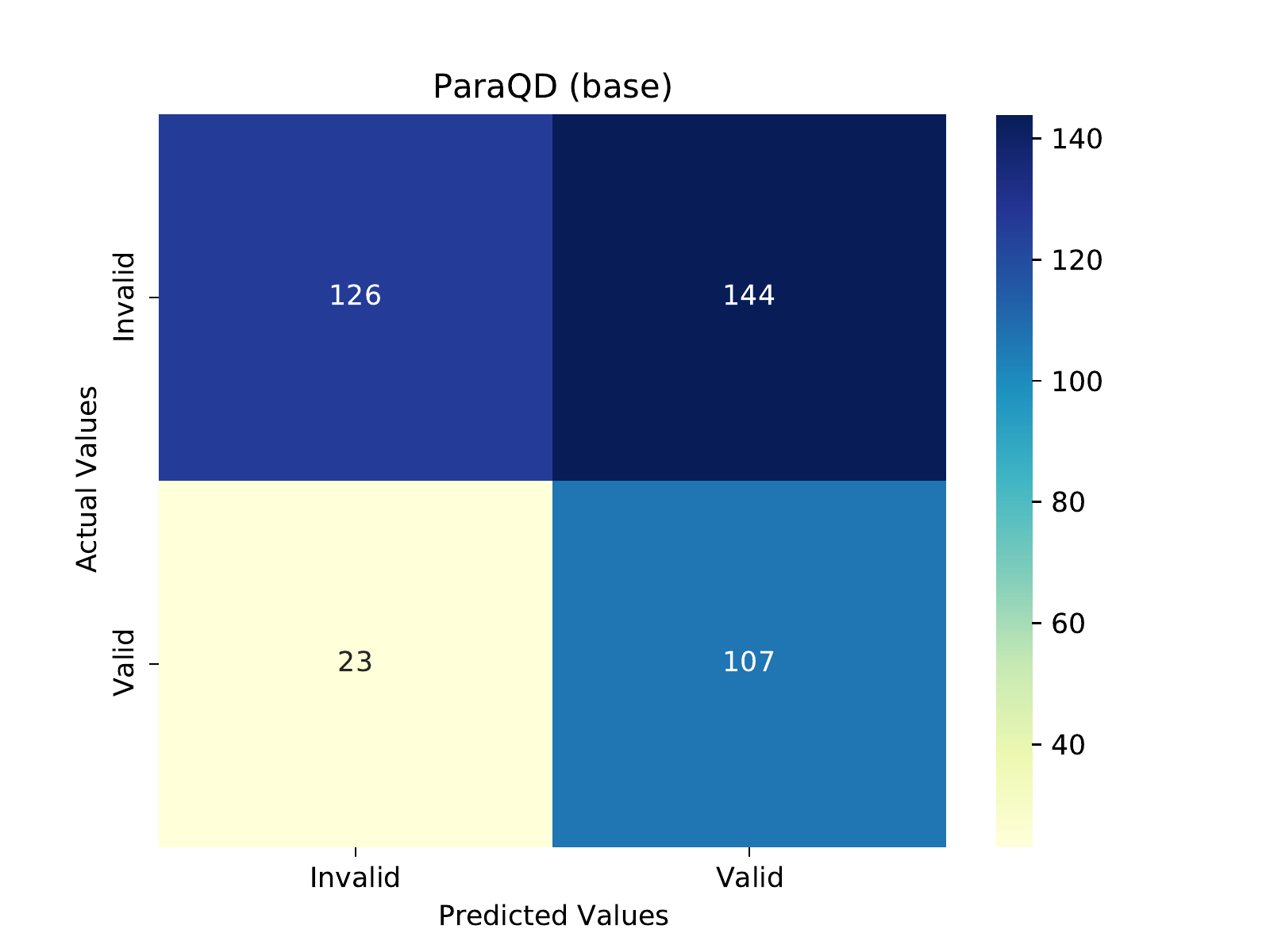}
\caption{ParaQD}
\end{subfigure}
% \hspace{0.3em}
\begin{subfigure}{0.32\linewidth}
%\vspace{-10pt}
\includegraphics[width=1.5\linewidth]{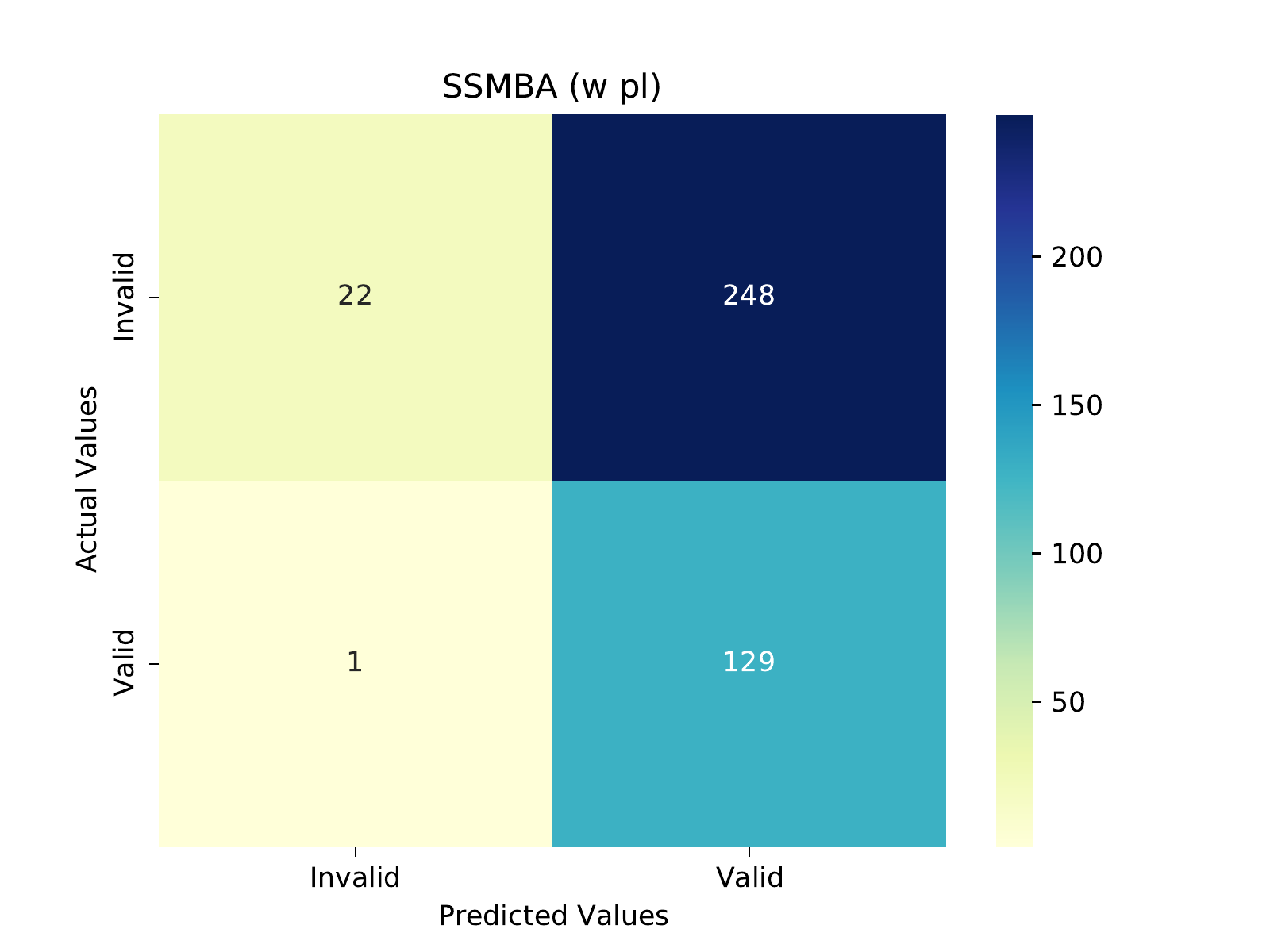}
\caption{SSMBA with pl}
\end{subfigure}%
\hspace{1em}
\begin{subfigure}{0.32\linewidth}
%\vspace{-10pt}
\includegraphics[width=1.5\linewidth]{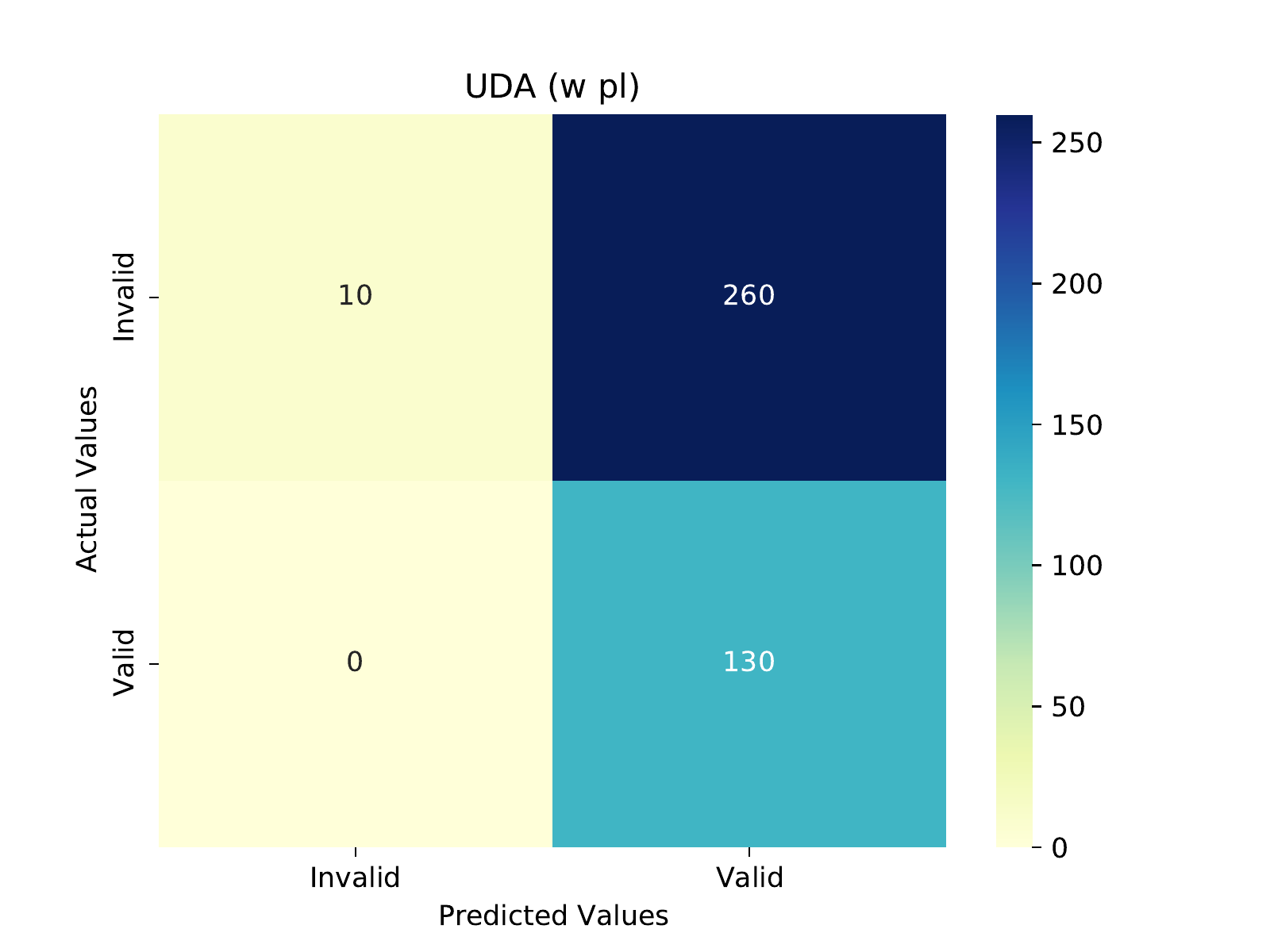}
\caption{UDA with pl}
\end{subfigure}
\label{subfig}
\begin{subfigure}{0.32\linewidth}
%\vspace{-18pt}
\includegraphics[width=1.5\linewidth]{images/confusion_matrices/EM/Pretrained.pdf}
\caption{Pretrained}
\end{subfigure}%
\begin{subfigure}{0.31\linewidth}
%\vspace{-9pt}
\includegraphics[width=1.5\linewidth]{images/confusion_matrices/EM/SSMBA__wo_pl_.pdf}
\caption{SSMBA}
\end{subfigure}
% \hspace{0.6em}
\begin{subfigure}{0.32\linewidth}
%\vspace{-18pt}
\includegraphics[width=1.5\linewidth]{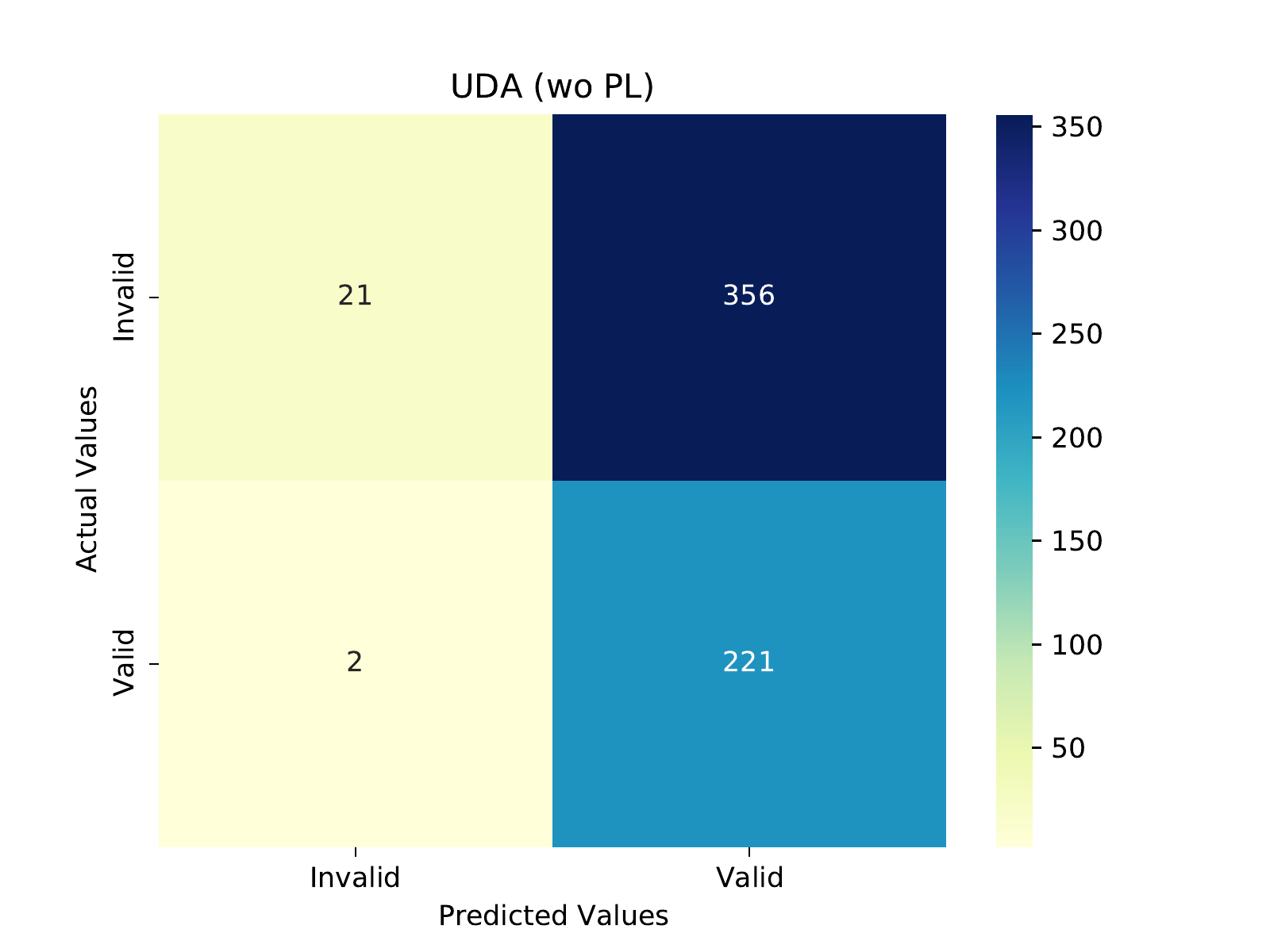}
\caption{UDA}
\end{subfigure}%
\hspace{1em}
\caption{Confusion matrices for all methods on SAWP.}
\label{fig:cm-SAWP}
\vspace{-2mm}
\end{figure*}

\begin{figure*}[hbt!]
% \centering
% \hspace{-2pt}
\begin{subfigure}{0.32\linewidth}
%\vspace{-1pt}
\hspace{-2pt}
\includegraphics[width=1.5\linewidth]{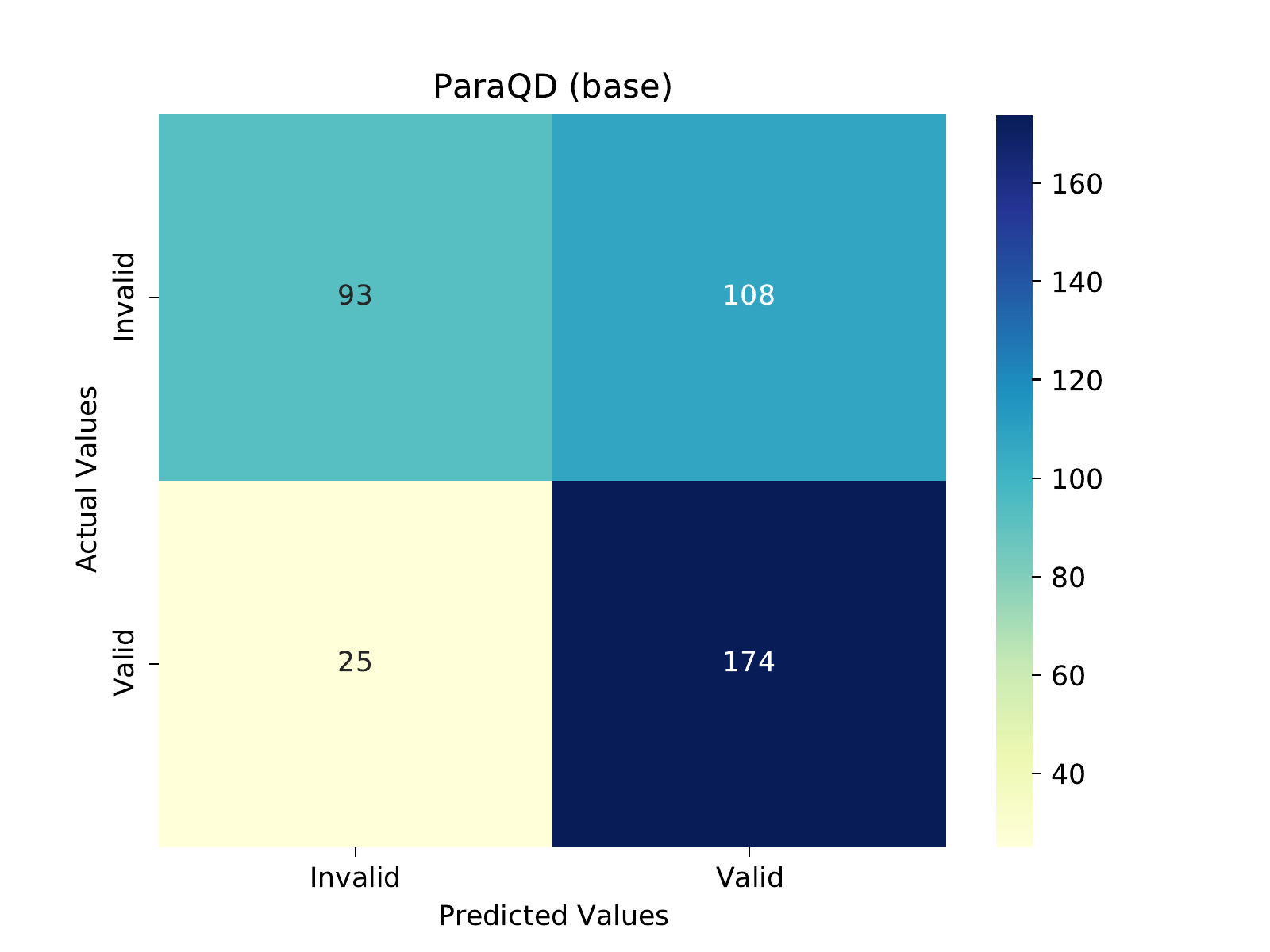}
\caption{ParaQD}
\end{subfigure}
% \hspace{0.3em}
\begin{subfigure}{0.32\linewidth}
%\vspace{-10pt}
\includegraphics[width=1.5\linewidth]{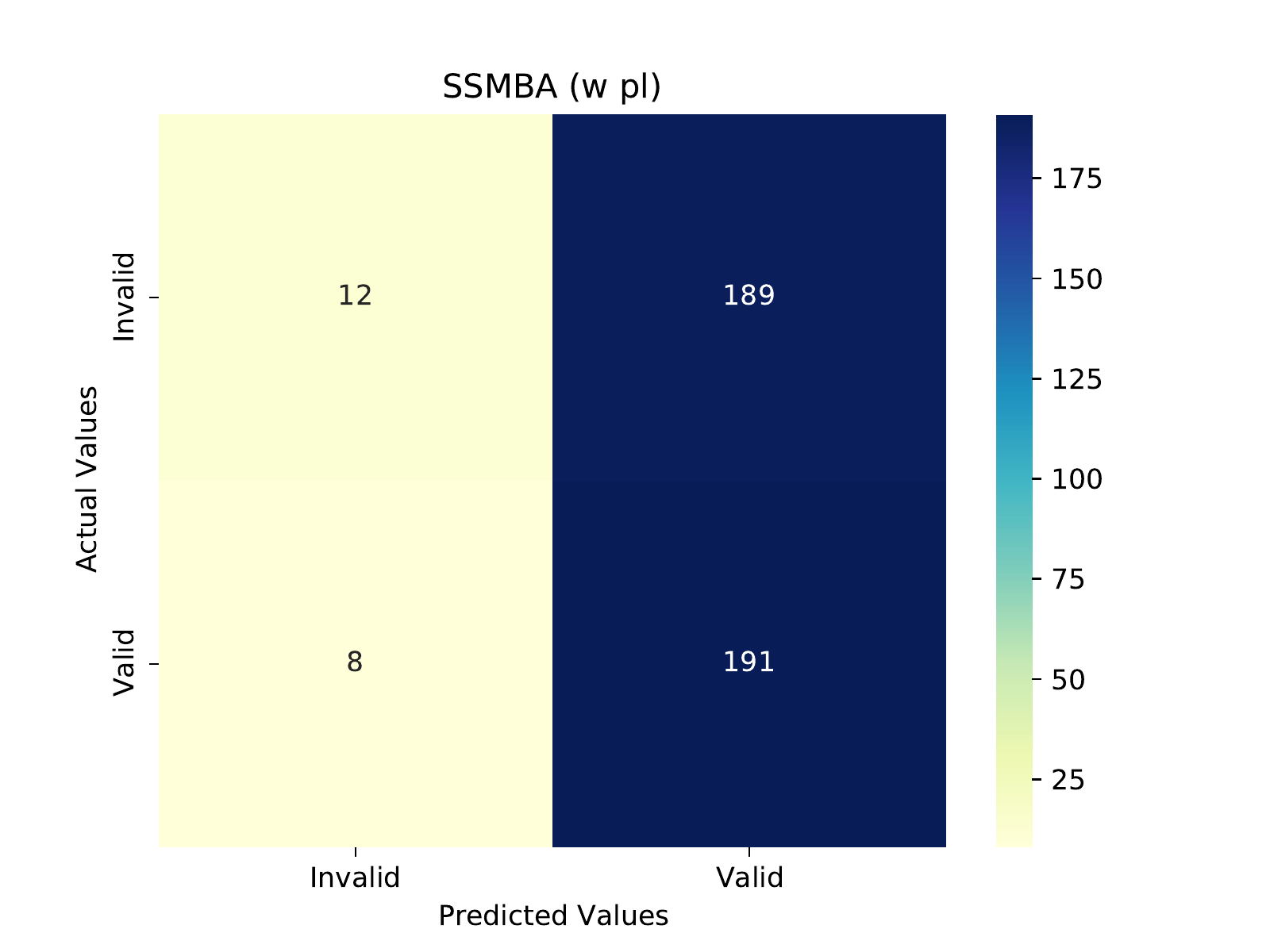}
\caption{SSMBA with pl}
\end{subfigure}%
\hspace{1em}
\begin{subfigure}{0.32\linewidth}
%\vspace{-10pt}
\includegraphics[width=1.5\linewidth]{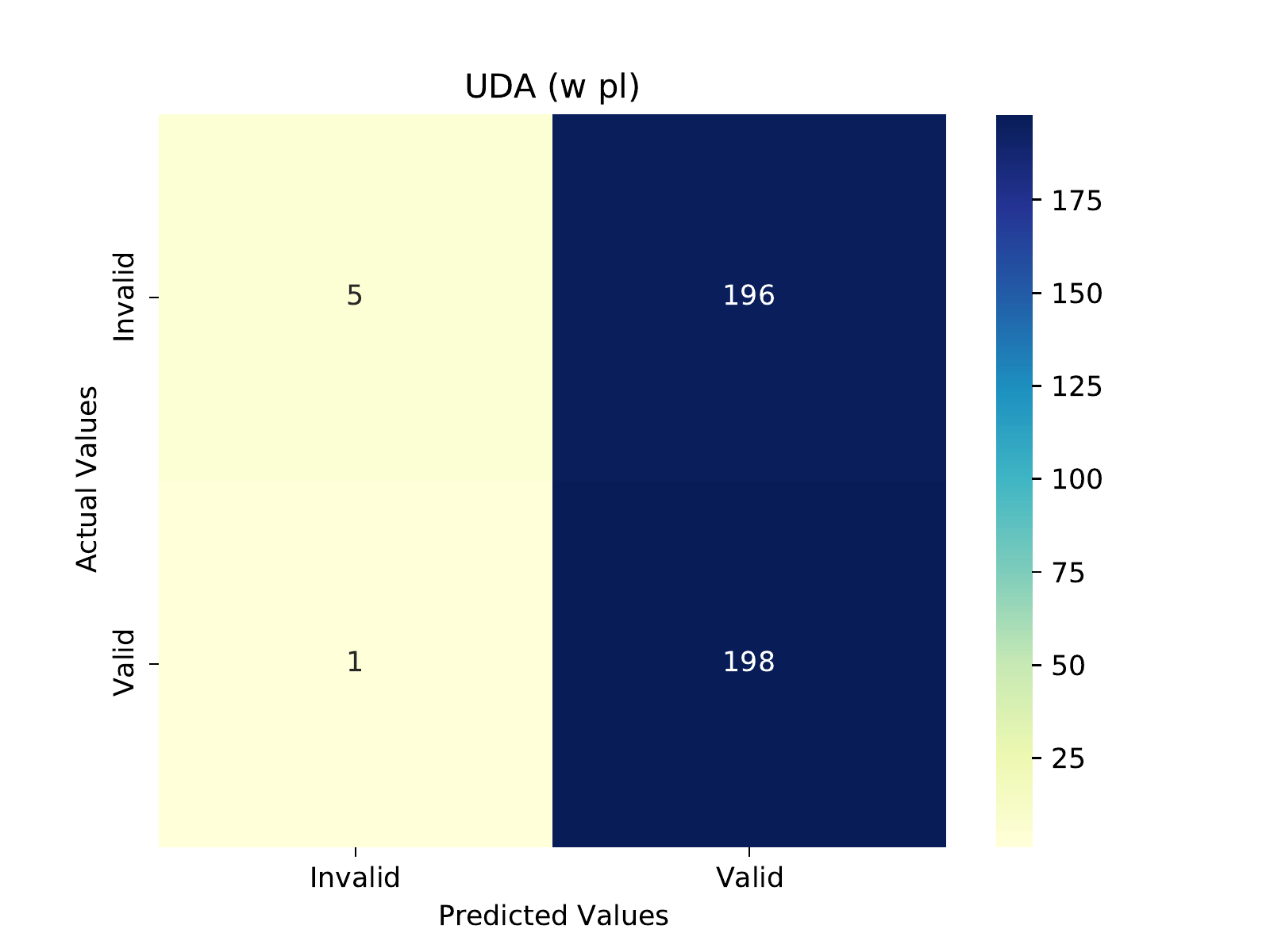}
\caption{UDA with pl}
\end{subfigure}
\label{subfig}
\begin{subfigure}{0.32\linewidth}
%\vspace{-18pt}
\includegraphics[width=1.5\linewidth]{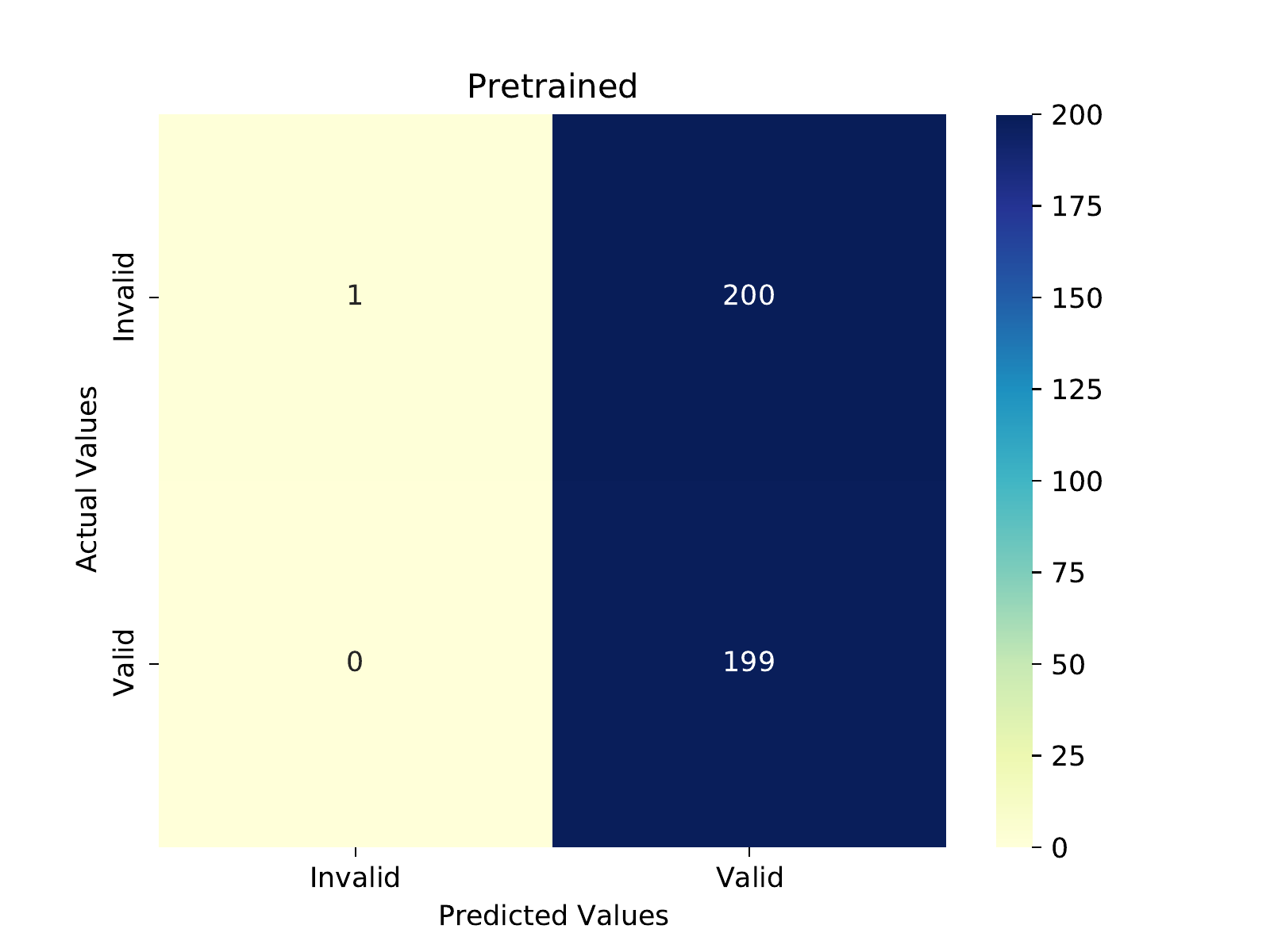}
\caption{Pretrained}
\end{subfigure}%
\begin{subfigure}{0.31\linewidth}
%\vspace{-9pt}
\includegraphics[width=1.5\linewidth]{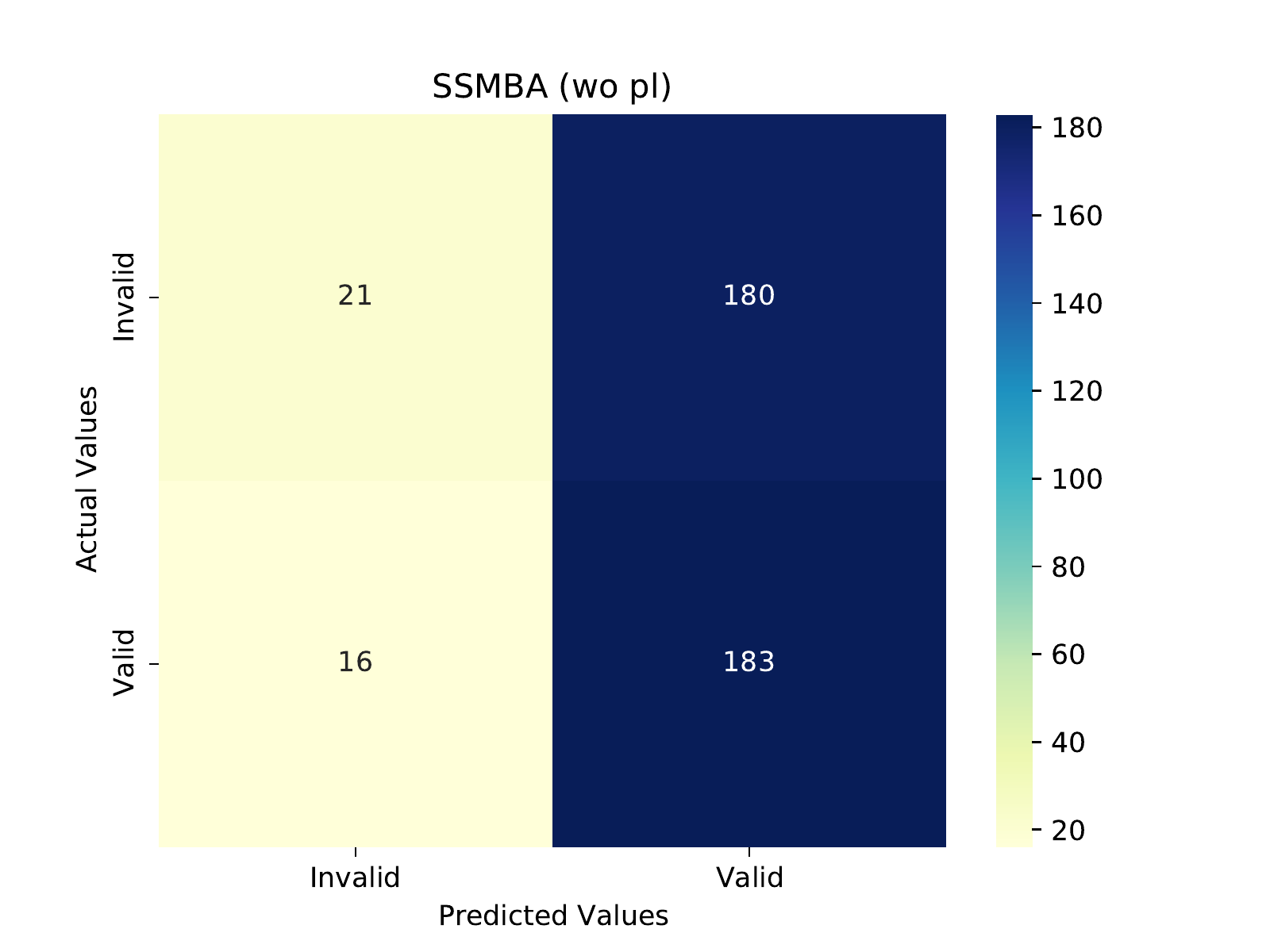}
\caption{SSMBA}
\end{subfigure}
% \hspace{0.6em}
\begin{subfigure}{0.32\linewidth}
%\vspace{-18pt}
\includegraphics[width=1.5\linewidth]{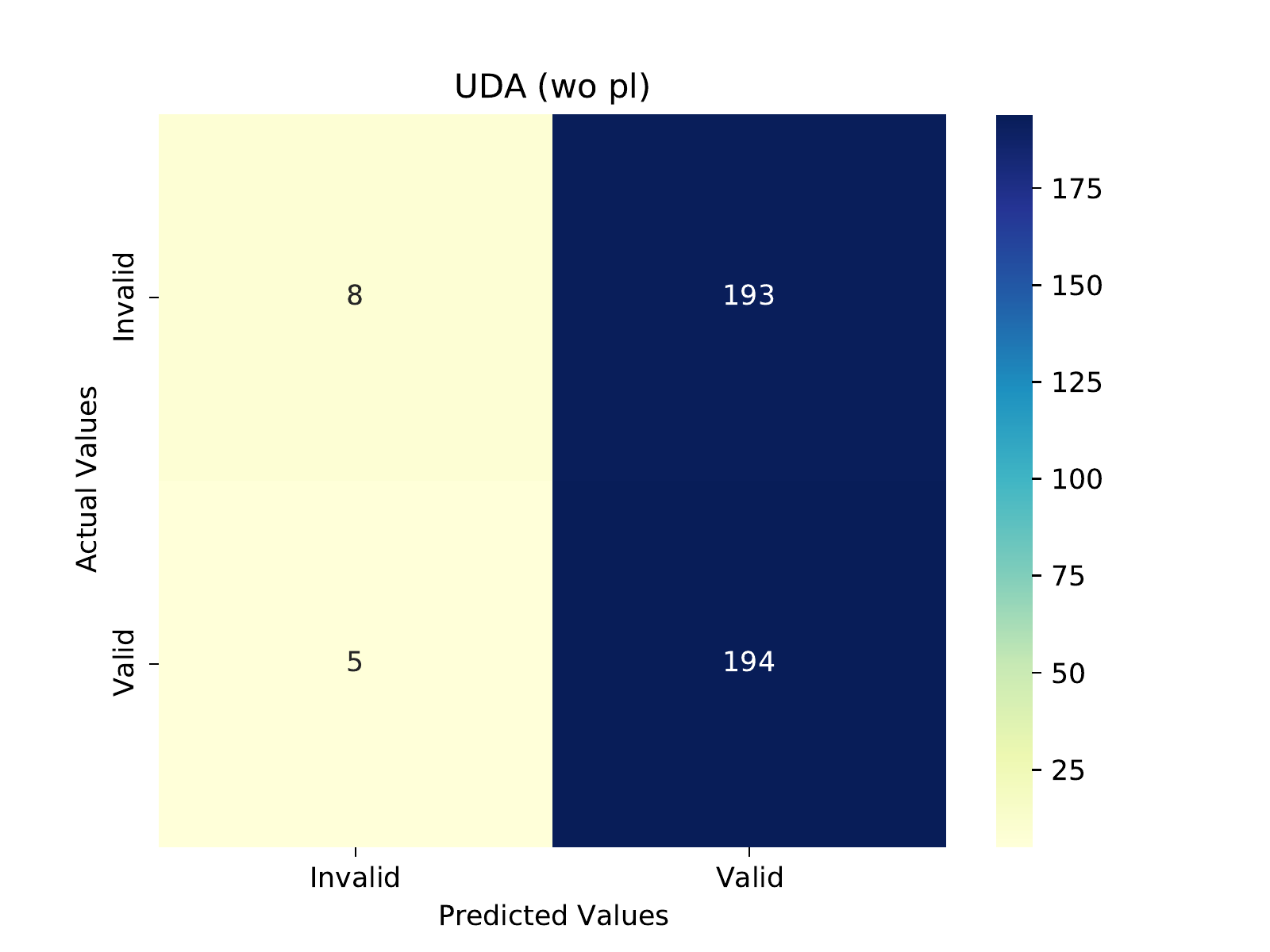}
\caption{UDA}
\end{subfigure}%
\hspace{1em}
\caption{Confusion matrices for all methods on PAWP}
\label{fig:cm-PAWP}
\vspace{-2mm}
\end{figure*}

\section{Augmentation: Further Details}
\label{appendix:augs}

\subsection{$f_1:$ Backtranslation}
We used WMT'19 FSMT \cite{ng2019facebook} \textit{en-de} and \textit{de-en} translation models, with language $A$ being $English$ and $B$ being $German$.
We used diverse beam search \cite{vijayakumar2018diverse} for decoding in the reverse translation step to introduce diversity and chose the candidate paraphrase with the maximum Levenshtein distance. This ensures that the model learns to give a high score for diverse paraphrases that retain critical information.

\subsection{$f_5$: Most Important Phrase Deletion}
To model $\Psi$, we use TopicRank \cite{bougouin-etal-2013-topicrank}. The methodology to select the most critical phrase is inspired by TSDAE \cite{wang2021tsdae}.

\subsection{$f_b$: Corrupted Sentence Reconstruction}
When corrupting the input sentence, we preserve the numbers, units and the last three tokens. This is done because if we corrupt the numbers or the units, the model cannot accurately reconstruct them and will replace them with random numbers and units. We preserve the last three tokens because corrupting them might lead the model to change the question as the last three tokens in a word problem are generally indicative of the question.

% \usepackage{multirow}

\section{Baselines: Pseudo Labelling}
\label{appendix:pl}
We use the same pretrained encoder (MiniLM) to first pseudo-label the samples (without being trained) and then train it using the pseudo-labelled samples. More formally, given an input $\Q$ and a paraphrase $\Q'$, we use the encoder to determine whether $\Q'$ is a positive or negative paraphrase of $\Q$ as follows:
  \begin{align*}
    &\rho_i, \zeta_i = ENC(\Q), ENC(\Q') \\
    &\lambda(\Q, \Q')=
    \begin{cases}
      1 &if\;cossim(\rho_i, \zeta_i) > \iota \\
      0 &if\;cossim(\rho_i, \zeta_i) \leq \iota 
     \end{cases}
  \end{align*}
where $\iota$ is the threshold for the cosine similarity, which we set to 0.8.

We observe that on AquaRAT, the performance decreases for both UDA and SSMBA due to pseudo labelling, while it increases on EM\_Math dataset. This can be due to the much higher percentage of pseudo-labelled negative samples for EM\_Math as shown in Table \ref{table:pl}, thus providing more information about detecting invalid paraphrases as seen in Figure \ref{fig:cm-EM}.

\section{Embedding Plots}
\label{appendix:plots}
We plotted the embeddings across triplets in the test set to observe the separation margin. The colour represents a triplet, while the symbol represents which component of the triplet it is (anchor, positive, negative). The left embedding plot is for 7 randomly chosen triplets in the data (to closely visualize the distances), while the one on the right is for all triplets (93).

\begin{table*}[htb!]
\caption{Effect of the Loss Function for ParaQD (on AquaRAT)}
\begin{tabular}{crrrrrrllp{0.6cm}}
\hline
\multirow{2}{*}{\textbf{Loss}} & \multicolumn{3}{c}{\textbf{Macro}} & \multicolumn{3}{c}{\textbf{Weighted}} & \multirow{2}{*}{\textbf{$\mu^+$}} & \multirow{2}{*}{\textbf{$\mu^-$}} & \multirow{2}{*}{\textbf{$\mu^s$}} \\ \cline{2-7}
 & \multicolumn{1}{l}{\textbf{P}} & \multicolumn{1}{l}{\textbf{R}} & \multicolumn{1}{l}{\textbf{F1}} & \multicolumn{1}{l}{\textbf{P}} & \multicolumn{1}{l}{\textbf{R}} & \multicolumn{1}{l}{\textbf{F1}} &  &  &  \\ \hline
Triplet & 0.678 & 0.695 & 0.687 & 0.762 & 0.625 & 0.687 & \multicolumn{1}{r}{0.77} & \multicolumn{1}{r}{-0.01} & \multicolumn{1}{r}{\textbf{0.78}} \\ \hline
MultipleNegativeRankingLoss & 0.708 & 0.716 & \textbf{0.712} & 0.801 & 0.627 & \textbf{0.704} & \multicolumn{1}{r}{0.89} & \multicolumn{1}{r}{0.474} & \multicolumn{1}{r}{0.416} \\ \hline
\end{tabular}
\label{table:loss}
\end{table*}

% \usepackage{multirow}
% \usepackage{multirow}

% \usepackage{multirow}
\begin{table*}[htb!]
\caption{An ablative analysis of all methods for different seeds on AquaRAT.}
\begin{tabular}{llrrrrrrrrp{0.6cm}}
\hline
\multirow{2}{*}{\textbf{Seed}} &
  \multirow{2}{*}{\textbf{Method}} &
  \multicolumn{3}{c}{\textbf{Macro}} &
  \multicolumn{3}{c}{\textbf{Weighted}} &
  \multicolumn{1}{l}{\multirow{2}{*}{\textbf{$\mu^+$}}} &
  \multicolumn{1}{l}{\multirow{2}{*}{\textbf{$\mu^-$}}} &
  \multicolumn{1}{l}{\multirow{2}{*}{\textbf{$\mu^s$}}} \\ \cline{3-8}
 &
   &
  \multicolumn{1}{l}{\textbf{P}} &
  \multicolumn{1}{l}{\textbf{R}} &
  \multicolumn{1}{l}{\textbf{F1}} &
  \multicolumn{1}{l}{\textbf{P}} &
  \multicolumn{1}{l}{\textbf{R}} &
  \multicolumn{1}{l}{\textbf{F1}} &
  \multicolumn{1}{l}{} &
  \multicolumn{1}{l}{} &
  \multicolumn{1}{l}{} \\ \hline
\multicolumn{1}{r}{\multirow{3}{*}{3407}} & ParaQD & 0.678 & 0.695 & \textbf{0.687} & 0.762 & 0.625 & \textbf{0.687} & 0.77  & -0.01 & \textbf{0.78}  \\
\multicolumn{1}{r}{}                      & UDA    & 0.661 & 0.512 & 0.577          & 0.786 & 0.332 & 0.467          & 0.995 & 0.966 & 0.029          \\  
\multicolumn{1}{r}{}                      & SSMBA  & 0.645 & 0.554 & 0.596          & 0.757 & 0.395 & 0.52           & 0.965 & 0.829 & 0.137          \\ \hline
\multirow{3}{*}{Seed Search}              & ParaQD & 0.684 & 0.694 & \textbf{0.689} & 0.772 & 0.614 & \textbf{0.684} & 0.828 & 0.055 & \textbf{0.772} \\
                                          & UDA    & 0.659 & 0.503 & 0.571          & 0.784 & 0.32  & 0.455          & 0.998 & 0.985 & 0.013          \\  
                                          & SSMBA  & 0.634 & 0.552 & 0.59           & 0.742 & 0.395 & 0.516          & 0.957 & 0.833 & 0.124          \\ \hline
\end{tabular}

\label{table:seed}
\end{table*}

\section{Loss Functions}
\label{appendix:loss}
Multiple Negatives Ranking Loss \footnotemark is a loss function, which, for anchor $a_i$ in the triplet $(a_i, p_i, n_i)$ considers $p_i$ as a positive sample and all $p_j$ and $n_k$ in the batch (such that $j!=i$) as negatives. It works by maximizing the log-likelihood of the softmax scores. The equation is similar to the one in \cite{henderson2017efficient}. 

\note{\url{https://www.sbert.net/docs/package\_reference/losses.html\#multiplenegativesrankingloss}}

\section{Performance on test set with training operators}
\label{appendix:trainops}
We also evaluated our method by generating the test set of AquaRAT using training operators. The results are present in Table \ref{table:trainops}. ParaQD outperforms the baselines by a significant margin across all metrics. However, this test set is not representative of real data distribution as it is suited for our method. Thus, we report the results only for the sake of completion and they should not be taken as representative.

\begin{table*}[htb!]
\caption{Performance of all methods on the test set of AquaRAT created using train operators.}
\begin{tabular}{lrrrrrrrrp{0.6cm}}
\hline
\multirow{2}{*}{\textbf{Method}} &
  \multicolumn{3}{c}{\textbf{Macro}} &
  \multicolumn{3}{c}{\textbf{Weighted}} &
  \multicolumn{1}{l}{\multirow{2}{*}{\textbf{$\mu^+$}}} &
  \multicolumn{1}{l}{\multirow{2}{*}{\textbf{$\mu^-$}}} &
  \multicolumn{1}{l}{\multirow{2}{*}{\textbf{$\mu^s$}}} \\ \cline{2-7}
 &
  \multicolumn{1}{l}{\textbf{P}} &
  \multicolumn{1}{l}{\textbf{R}} &
  \multicolumn{1}{l}{\textbf{F1}} &
  \multicolumn{1}{l}{\textbf{P}} &
  \multicolumn{1}{l}{\textbf{R}} &
  \multicolumn{1}{l}{\textbf{F1}} &
  \multicolumn{1}{l}{} &
  \multicolumn{1}{l}{} &
  \multicolumn{1}{l}{} \\ \hline
Pretrained    & 0.25  & 0.5   & 0.333          & 0.25  & 0.5            & 0.333 & 0.966 & 0.92   & 0.046          \\ 
UDA           & 0.626 & 0.505 & 0.559          & 0.626 & 0.505          & 0.559 & 0.995 & 0.973  & 0.022          \\ 
UDA (w pl)    & 0.681 & 0.511 & 0.584          & 0.681 & 0.511          & 0.584 & 0.99  & 0.965  & 0.025          \\ 
SSMBA         & 0.667 & 0.532 & 0.592          & 0.667 & 0.532          & 0.592 & 0.965 & 0.871  & 0.094          \\ 
SSMBA (w pl)  & 0.705 & 0.518 & 0.597          & 0.705 & 0.518          & 0.597 & 0.987 & 0.927  & 0.06           \\ 
ParaQD (ours) & 0.903 & 0.895 & \textbf{0.899} & 0.903 & 0.895 & \textbf{0.899} & 0.927 & -0.656 & \textbf{1.583} \\ \hline
\end{tabular}

\label{table:trainops}
\end{table*}

\begin{table*}[htb!]
\caption{An ablative analysis of all methods for 3 different encoders on AquaRAT. We observe that regardless of the encoder used, we outperform the baselines on all metrics.}
\hspace{-1cm}
\begin{tabular}{llrrrrrrrrp{0.6cm}}
\hline
\multirow{2}{*}{\textbf{Encoder}} &
  \multirow{2}{*}{\textbf{Method}} &
  \multicolumn{3}{c}{\textbf{Macro}} &
  \multicolumn{3}{c}{\textbf{Weighted}} &
  \multicolumn{1}{l}{\multirow{2}{*}{\textbf{$\mu^+$}}} &
  \multicolumn{1}{l}{\multirow{2}{*}{\textbf{$\mu^-$}}} &
  \multicolumn{1}{l}{\multirow{2}{*}{\textbf{$\mu^s$}}} \\ \cline{3-8}
 &
   &
  \multicolumn{1}{l}{\textbf{P}} &
  \multicolumn{1}{l}{\textbf{R}} &
  \multicolumn{1}{l}{\textbf{F1}} &
  \multicolumn{1}{l}{\textbf{P}} &
  \multicolumn{1}{l}{\textbf{R}} &
  \multicolumn{1}{l}{\textbf{F1}} &
  \multicolumn{1}{l}{} &
  \multicolumn{1}{l}{} &
  \multicolumn{1}{l}{} \\ \hline
\multirow{3}{*}{\textbf{all-minilm-L12-v1 (base)}} & ParaQD & 0.678 & 0.695 & \textbf{0.687} & 0.762 & 0.625 & \textbf{0.687} & 0.77  & -0.01 & \textbf{0.78}  \\
                                                   & UDA    & 0.661 & 0.512 & 0.577          & 0.786 & 0.332 & 0.467          & 0.995 & 0.966 & 0.029          \\ 
                                                   & SSMBA  & 0.645 & 0.554 & 0.596          & 0.757 & 0.395 & 0.52           & 0.965 & 0.829 & 0.137          \\ \hline
\multirow{3}{*}{\textbf{MPNet}}                    & ParaQD & 0.703 & 0.726 & \textbf{0.714} & 0.785 & 0.659 & \textbf{0.717} & 0.858 & 0.201 & \textbf{0.656} \\
                                                   & UDA    & 0.659 & 0.503 & 0.571          & 0.784 & 0.32  & 0.455          & 0.99  & 0.953 & 0.037          \\ 
                                                   & SSMBA  & 0.66  & 0.508 & 0.574          & 0.785 & 0.327 & 0.462          & 0.985 & 0.94  & 0.045          \\ \hline
\multirow{3}{*}{\textbf{all-minilm-L6-v2}}         & ParaQD & 0.671 & 0.679 & \textbf{0.675} & 0.758 & 0.598 & \textbf{0.668} & 0.799 & 0.083 & \textbf{0.716} \\
                                                   & UDA    & 0.659 & 0.503 & 0.571          & 0.784 & 0.32  & 0.455          & 0.994 & 0.979 & 0.015          \\ 
                                                   & SSMBA  & 0.661 & 0.513 & 0.578          & 0.786 & 0.334 & 0.469          & 0.992 & 0.941 & 0.051          \\ \hline
\end{tabular}

\label{table:encoder}
\end{table*}

\begin{table*}[htb!]
\centering
\caption{Pseudo Labelling Statistics. Positive\% represents the percentage of total samples pseudo-labelled as positive, while Negative\% represents the percentage of total samples pseudo-labelled as negatives.}
\begin{tabular}{llrp{0.6cm}}
\hline
Dataset                   & Method       & \multicolumn{1}{l}{Positive \%} & \multicolumn{1}{l}{Negative \%} \\ \hline
\multirow{2}{*}{AquaRAT}  & UDA (w pl)   & 87.29                           & 12.71                           \\
                          & SSMBA (w pl) & 75.69                           & 24.31                           \\ \hline
\multirow{2}{*}{EM\_Math} & UDA (w pl)   & 72.16                           & 27.84                           \\ \cline{2-4} 
                          & SSMBA (w pl) & 55.09                           & 44.91                           \\ \hline
\end{tabular}

\label{table:pl}
\end{table*}

\section{Encoder Ablations and Seed Optimization}
\label{appendix:encoder}
The results of varying the encoder are shown in Table \ref{table:encoder}. We vary the encoder and experiment with MiniLM (12 layers), MiniLM (6 layers), and MPNet \cite{song2020mpnet}. We choose the encoders considering different metrics such as average performance on semantic search and encoding speed ($\# \ of sentences/sec$)\footnote{The metrics are obtained from \url{https://www.sbert.net/docs/pretrained\_models.html}}. For instance MPNet can encode about \textit{2500} sentences/sec whereas MiniLM (12 layers) (all-minilm-L12-v1) can encode about 7500 sentences/sec and MiniLM (6 layers) (all-minilm-L6-v2) can encode about 14200 sentences/sec. Also the average performance on semantic search benchmarks is of order $all-minilm-L12-v1$ $>$ $all-minilm-L6-v2$ $>$ MPNet. In our setting, from table \ref{table:encoder} we can observe that MiniLM (12 layers) surpasses the other encoders as measured by the separation metric ($\mu^s$). However, when we observe Macro and weighted F1 scores, MPNet surpasses the other two encoders. Since we are more concerned about how positive and negative paraphrases are separated in the vector space, we choose MiniLM (12 layers) (all-minilm-L12-v1) for all our main experiments, as shown in Table 1. We can also observe that the proposed method (ParaQD) outperforms all other baselines. This demonstrates the robustness of the proposed augmentation method and shows the performance gain when compared to other baselines is invariant to changes in encoders.
We also vary the seed values to check for the robustness of our method. We compare the random seed value (3407) with the seed optimization method proposed in the Augmented SBERT paper \cite{thakur2021augmented}. For seed optimization, we search for the best seed in the range [0-4] as recommended in the original work by training 20\% of the data and comparing the results on the validation set. We then select the best performing seed and train using that particular seed. The results of the experiments are shown in Table \ref{table:seed}. We observe that the best seed obtained through seed optimization and the seed value of 3407 nearly yield similar performance. We also observe that the proposed method outperforms the baselines demonstrating the robustness of the proposed method to seed randomization.

\end{document}